\documentclass[journal]{IEEEtran}
\usepackage{amsmath,amsfonts}
\usepackage{algorithmic}
\usepackage{algorithm}
\usepackage{array}
\usepackage{textcomp}
\usepackage{xcolor}
\usepackage{stfloats}
\usepackage{url}
\usepackage{verbatim}
\usepackage{graphicx}
\usepackage[backend=biber,style=ieee]{biblatex}
\usepackage{dsfont}
\usepackage{multicol}
\usepackage{multirow}
\usepackage{blindtext}
\usepackage{subfigure}
\begin{document}
% RoboFiSense: 
\title{RoboFiSense: Attention-Based Robotic Arm Activity Recognition with WiFi Sensing}

\author{Rojin Zandi$^\dagger$, Kian Behzad$^\dagger$, Elaheh Motamedi$^\dagger$, Hojjat Salehinejad$^\star$$^\ddagger$, \textit{Senior Member, IEEE}, and \\ Milad Siami$^\ddagger$$^\dagger$ \textit{Senior Member, IEEE}
\thanks{
This material is based upon work supported in part at Northeastern University by grants ONR N00014-21-1-2431, NSF 2121121, the U.S. Department of Homeland Security under Grant Award Number 22STESE00001-01-00, and by the Army Research Laboratory under Cooperative Agreement Number W911NF-22-2-0001. The views and conclusions contained in this document are solely those of the authors and should not be interpreted as representing the official policies, either expressed or implied, of the U.S. Department of Homeland Security, the Army Research Office, or the U.S. Government.}
\thanks{$^\dagger$Department of Electrical \& Computer Engineering, Northeastern University, Boston, MA, USA  { (e-mails: {\tt\small \{zandi.r, behzad.k, Motamedi.e, m.siami\}@northeastern.edu}).}}
\thanks{
$^\star$Kern Center for the Science of Health Care Delivery and Department of Artificial Intelligence and Informatics, Mayo Clinic, Rochester, MN, USA {(e-mail: {\tt\small salehinejad.hojjat@mayo.edu}).}}
\thanks{$^\ddagger$Share senior authorship.}}

% The paper headers
% \markboth{Journal of Selected Topics in Signal Processing,~Vol.~xx, No.~xx, xxx~xxxx}%
% {Shell \MakeLowercase{\textit{et al.}}: A Sample Article Using IEEEtran.cls for IEEE Journals}

\maketitle

\begin{abstract}
Despite the current surge of interest in autonomous robotic systems, robot activity recognition within restricted indoor environments remains a formidable challenge. Conventional methods for detecting and recognizing robotic arms' activities often rely on vision-based or light detection and ranging (LiDAR) sensors, which require line-of-sight (LoS) access and may raise privacy concerns, for example, in nursing facilities. 
This research pioneers an innovative approach harnessing channel state information (CSI) measured from WiFi signals, subtly influenced by the activity of robotic arms. We developed an attention-based network to classify eight distinct activities performed by a Franka Emika robotic arm in different situations. Our proposed bidirectional vision transformer-concatenated (BiVTC) methodology aspires to predict robotic arm activities accurately, even when trained on activities with different velocities, all without dependency on external or internal sensors or visual aids. Considering the high dependency of CSI data on the environment motivated us to study the problem of sniffer location selection, by systematically changing the sniffer's location and collecting different sets of data. Finally, this paper also marks the first publication of the CSI data of eight distinct robotic arm activities, collectively referred to as RoboFiSense. This initiative aims to provide a benchmark dataset and baselines to the research community, fostering advancements in the field of robotics sensing.
\end{abstract}

\section{Introduction}
\IEEEPARstart{I}{n} recent years, the spotlight in technology has been directed toward the growing field of autonomous robotic systems, powered by remarkable advancements in artificial intelligence. These systems have garnered considerable attention due to their remarkable capability to function autonomously across diverse environments, independent of human intervention \cite{atkeson2015no, haddadin2017robot, de2008atlas}. Autonomous robots find practical utility across a diverse range of industries, as they play pivotal roles in manufacturing, from executing precision welding tasks \cite{bischoff2010kuka} to aiding in environmental monitoring through applications like ocean exploration \cite{aracri2021soft}. In the field of healthcare, these robots navigate the intricate terrain of surgical procedures \cite{lanfranco2004robotic} and contribute to patient rehabilitation efforts \cite{atashzar2016characterization}. Their expertise is showcased in their ability to handle tasks that are either perilous or monotonous, execute them with an unrivaled degree of accuracy and precision.

Understanding a robot's activity in an environment is not only fundamental for safe and efficient operation but also vital for enhancing its utility across a range of applications~\cite{javaid2021substantial, motamedi2024robustness}. Nonetheless, amidst this remarkable technological progress, the challenge of accurately predicting the activity of these robots remains a formidable obstacle in the field of robotics.
Traditionally, robotic activity recognition has heavily relied on visual and spatial sensors such as cameras and light detection and ranging (LiDAR) systems. Cameras provide robots with a human-like vision, capturing the world in vivid detail. LiDAR, alternatively, offers a different perspective by creating precise 3D maps through laser pulses \cite{li2020building}.

The intricacies of modern robotic applications require solutions that transcend the limitations of vision and sensory-based methods. Challenges like low-light conditions \cite{petrlik2021lidar}, visual obstructions  \cite{correa2012mobile}, and environments with no line-of-sight (NLoS) have underscored the necessity for more versatile and adaptive sensing technologies \cite{yang2024mm}. Another pertinent concern associated with vision-based techniques is the intrusion into privacy, notably in the context of surveillance systems.

WiFi sensing is an emerging discipline that utilizes the widespread WiFi infrastructure for various applications such as human activity recognition (HAR)~\cite{salehinejad2022litehar,yousefi2017surv}, and presence detection \cite{9745814}. In this approach, changes in the environment where the WiFi is operating can be captured by analyzing the channel state information (CSI)~\cite{ma2019wifi} using novel machine learning techniques. One of the distinguishing features of WiFi sensing is its ability to overcome visual limitations. Unlike cameras, which depend on optical systems, or LiDAR sensors which rely on laser-based measurements, WiFi signals may not be impacted by noise such as lighting conditions~\cite{zandi2023robot}. In addition, WiFi sensing provides a privacy advantage. Unlike cameras, which capture visual information, WiFi sensing operates on radio frequency signals, making it a suitable option for situations where data security and privacy compliance are critical \cite{gu2017mosense}.

The integration of robots into everyday environments, especially in privacy-sensitive sectors like healthcare, necessitates non-intrusive monitoring methods such as CSI measurements \cite{9745814}. These enhance both privacy and system robustness in a multimodal setup with other sensors \cite{yang2024mm}. Robots enable large-scale data collection without ethical issues, aiding the development of specialized robotic activity recognition (RAR) models and advancing transfer learning. Given the unique characteristics of robot activities, specialized RAR models are essential, as a direct application of HAR algorithms may not yield optimal results, underscoring the need for tailored model evaluation.

\noindent \textbf{Our Contributions:} As WiFi sensing technology has evolved, its application in enhancing robotic systems, particularly in the area of activity recognition, has become increasingly prominent \cite{wang2023wi,zandi2023robot}. Leveraging the adaptability of CSI, this research delves into the comprehensive utilization of WiFi sensing for RAR across diverse activities and scenarios. The key contributions of this study include:

\begin{enumerate}
\item \textbf{Introducing a Benchmark Dataset for RAR with CSI Measurements:} Aiming to foster research in the field of RAR, we introduce RoboFiSense, the first inaugural public CSI dataset showcasing eight unique robotic arm activities\footnote{\url{https://github.com/SiamiLab/RoboFiSense}}, expanding our prior classification of four actions \cite{zandi2023robot}.

\item \textbf{Development of a Custom Vision Transformer Model for RAR:} Building on our previous convolutional neural network (CNN)-based endeavors \cite{zandi2023robot}, we introduce a vision transformer (ViT)-based model to identify eight distinct activities by a Franka Emika robotic arm, assessing the model's resilience against different activity velocities and the effectiveness of attention-based mechanisms in WiFi sensing \cite{9745814}.

\item \textbf{Analysis of Robotic Arm Velocity in RAR:} Acknowledging velocity variation challenges in HAR~\cite{hasanzadeh2023hand}, we present a novel evaluation scheme by systematically varying the robot's velocity across all activities during data collection, allowing a thorough analysis of our dataset under controlled velocity conditions to gauge machine learning model's efficacy.

\item \textbf{CSI Sampling Frequency Investigation:} In order to examine the role of sampling rate in CSI data collection, impact of sampling frequency rate on machine learning models performance across different velocities and sampling rates is investigated. 

% down-sampled the original $30$Hz CSI data to various frequencies, assessing the impact on model accuracy .

\item \textbf{Exploring Sniffer Positioning in RAR:} Addressing the critical factor of sniffer placement in WiFi sensing, a detailed examination is provided based on strategic sniffer deployments across a grid. 
% to understand impact of location influences CSI data collection quality and consistency.

\end{enumerate}

% These efforts significantly contribute to the advancement of WiFi-based robotic activity recognition, enhancing methodologies and data resources to meet the increasing need for versatile and secure sensing technologies.

The structure of this paper is organized as follows: Section \ref{sec: background} provides the foundational background, elucidating the essential concepts and related works. Section \ref{sec: prop method} details the proposed methods, describing the theoretical and practical aspects of our approach. In Section \ref{sec: dataset}, the data collection process and structuring of RoboFiSense dataset is discussed. Section \ref{sec:exp} delineates the experimental setup, detailing the methodologies utilized for data processing and analysis. It also presents the results and offers a thorough analysis, delving into the implications and insights gleaned from this study.

\section{Background} \label{sec: background}
% \subsection{Mathematical Notations} 
% Throughout the paper, we adopt standard mathematical notations to enhance clarity and readability. The sets of real and complex numbers are represented as $\mathbb{R}$ and $\mathbb{C}$, respectively. Vectors are denoted in lowercase letters (e.g., $\mathbf{x}$), while matrices are represented in uppercase bold letters (e.g., $\mathbf{H}$). Calligraphic letters denote sets (e.g., $\mathcal{S}$) and the transpose of matrix $\mathbf A$ is denoted by $\mathbf A^\top$.

\begin{figure}[!t]
    \centering
    \includegraphics[width=0.34\textwidth]{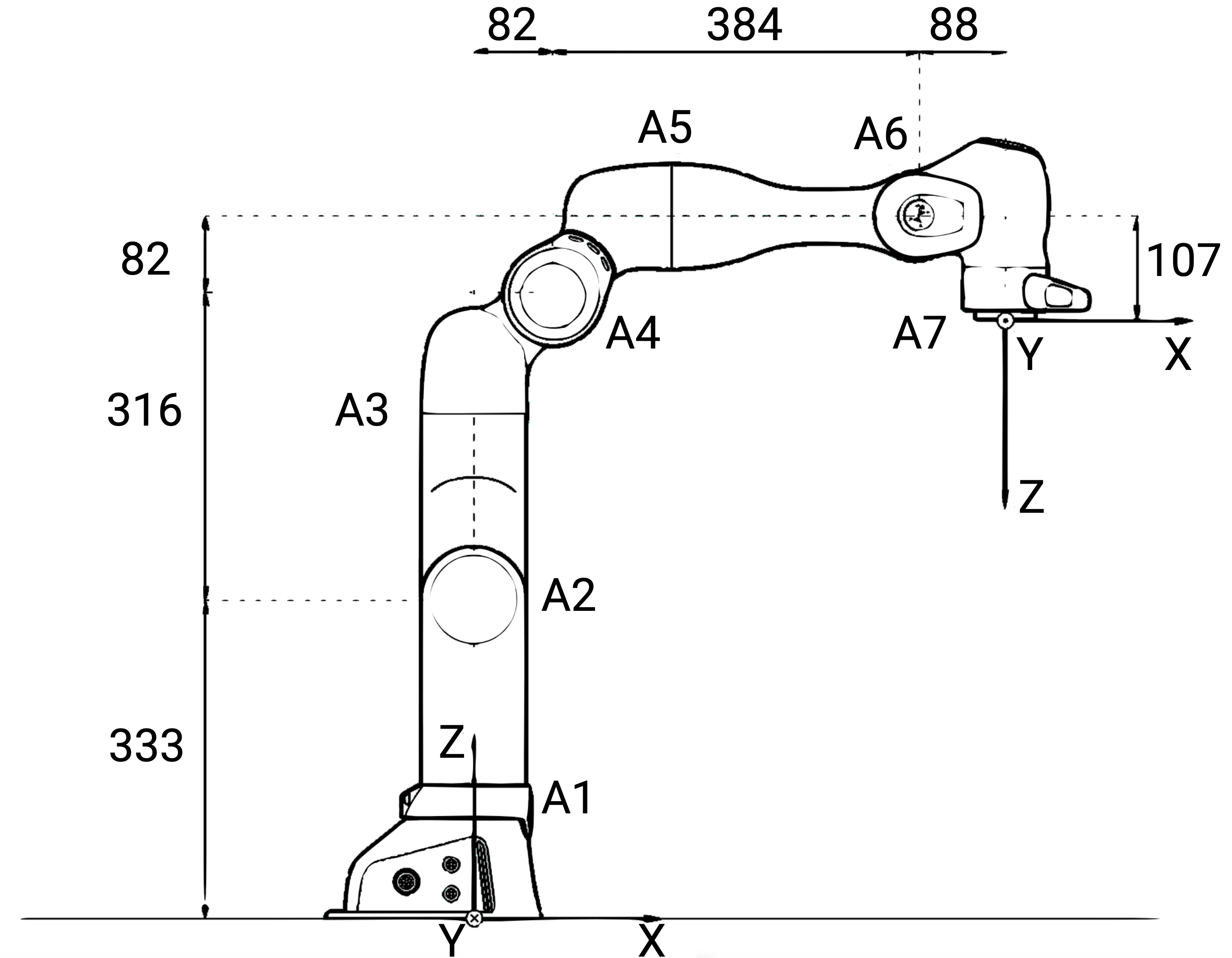}
    \caption{A Franka Emika robot with annotated joints and axis~\cite{frankaemika}.}
    \label{fig:FrankaEmika}
\end{figure}

\subsection{Franka Emika Robotic Arm}
% In this subsection, we present the adaptation and customization of demo activities on the Franka Emika robotic arm.
The Franka Emika robotic arm, as illustrated in Figure \ref{fig:FrankaEmika}, is a type of system known as a collaborative robot or cobot~\cite{frankaemika}. It can operate in industrial setups as well as right next to people, assisting them with tasks without posing a risk. Unlike typical factory robots, which are often put inside cages due to their potential danger, this arm can safely work alongside humans \cite{haddadin2022franka}. It is designed to perform tasks that require direct physical contact in a carefully controlled manner. These tasks include drilling, screwing, and buffing, as well as a variety of inspection and assembly tasks.

The Franka Emika robotic arm boasts a $7$-axis configuration, providing a three kg payload capacity and an impressive reach of $850$ mm. The robot weighs approximately $18$ kg and its repeatability is $0.1$ mm. Repeatability is a measure of the ability of the robot to consistently reach a specified point. The robot works as a torque-controlled robot, using strain gauges to measure forces on all of its seven joints. 

\begin{figure*}
    \centering
    \includegraphics[width=1\textwidth]{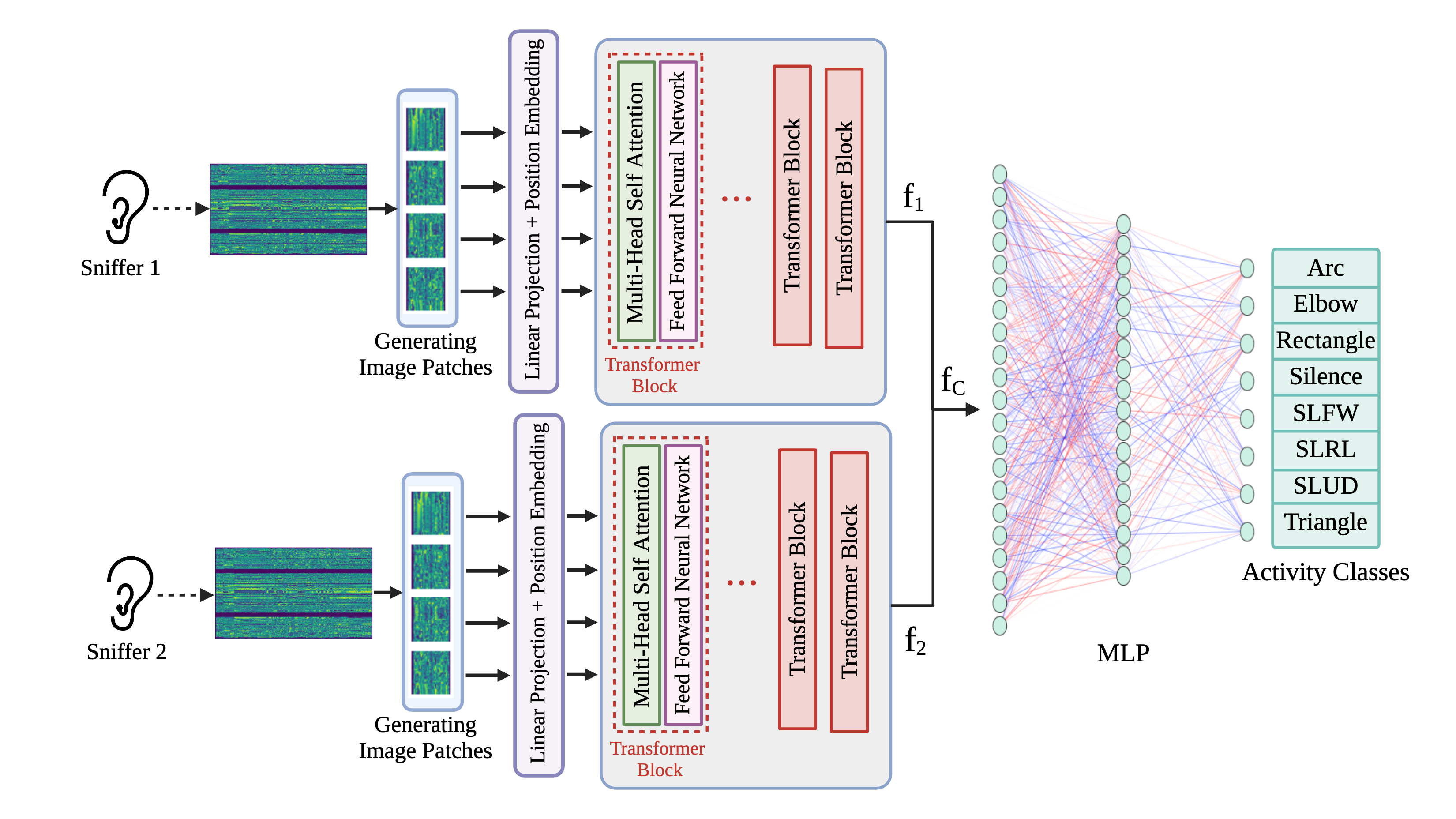}
    \caption{Architecture of the proposed bidirectional vision transformer-concatenated (BiVTC) model. The collected channel state information (CSI) measurements from each sniffer are separately patched, encoded, and fed to the transformer blocks for feature extraction. The feature vectors $\mathbf{f}_1$ and $\mathbf{f}_2$ are concatenated as $\mathbf{f}_c$ and passed as input to a multi-layer perceptron (MLP) network. }
    \label{fig:FE-Transforme}
\end{figure*}
% \vspace{-0.2cm}

\subsection{Channel State Information}
As wireless signals travel, they encounter obstacles in the environment, leading to reflections and scattering, which is also known as multipath fading \cite{yang2013rssi}. CSI enables the analysis of subcarrier propagation from the transmitter to the receiver in wireless communication \cite{wang2019survey}. The channel model can be expressed as
\begin{equation}
    \mathbf{y} = \mathbf{Hx} + \boldsymbol{\eta}, 
\end{equation}
where $\mathbf{x}$, $\mathbf{y}$ and $\boldsymbol{\eta}$ represent the transmitted signal vector, received signal vector, and additive noise vector, respectively~\cite{wang2019survey}. The channel matrix $\mathbf{H} \in \mathbb{C}^{T\times S}$ encapsulates the effects of the wireless channel, including multipath propagation, fading, and other impairments, defined as
\begin{equation}
    \mathbf{H}=\left[\begin{array}{cccc}h_1[1] & h_2[1] & \ldots & h_S[1] \\ h_1[2] & h_2[2] & \ldots & h_S[2] \\ \vdots & \vdots & \ddots & \vdots \\ h_{1}[T] & h_{2}[T] & \ldots & h_{S}[T]\end{array}\right],
\end{equation}
where $S$ and $T$ represent the number of subcarriers for each antenna and transmitted packets, respectively.
Each element of the matrix $\mathbf{H}$ corresponds to a complex value, denoted as the channel frequency response, defined as
\begin{equation}
    h_s[t] = a_s e^{j\phi_s},
\end{equation}
where $a_s$ and $\phi_s$ represent the amplitude and phase of subcarrier $s$, at timestamp $t$, respectively. For the purpose of HAR~\cite{salehinejad2022litehar,yousefi2017surv,widar2019} and RAR~\cite{zandi2023robot}, studies focus solely on $\mathbf{A}\in \mathbb{R}^{T\times S}$, which corresponds to the element-wise amplitude of $\mathbf{H}$, disregarding the phase component.

% The channel state information (CSI) is presented in the frequency domain, and to transition to the time domain, we can compute the inverse fast Fourier transform as follows
% \begin{equation}
%    \mathrm{h[n]} = \sum_{l=0}^{s-1} \mathbf{h_{l}} e^{-j\pi nl/s},
% \end{equation}
% where $\mathrm{h[n]}$ represents the channel impulse response (CIR), offering a time-domain representation of the channel's response to an impulse signal \cite{hernandez2022wifi}. The CIR describes how the transmitted signal propagates through the channel, arriving at the receiver with different time delays due to multipath effects.

\subsection{Advancements in WiFi Sensing for Robotics}

The remarkable success achieved by WiFi sensing in HAR has propelled its widespread adoption in the field of robotics~\cite{zegeye2016wifi,olivera2006wifi,zandi2023robot}. WiFi sensing has the distinct advantage of relying on existing infrastructure, which makes it highly cost-effective and suitable for indoor environments where GPS signals may be unreliable or unavailable \cite{7905633}. This innovative approach can enable robots to navigate and operate autonomously in complex and dynamic environments, such as warehouses, hospitals, and disaster-stricken areas \cite{rohrig2008tracking, sayed2020centralized}. 
WiFi technology has also found practical applications in RAR~\cite{zandi2023robot}. Machine learning has achieved remarkable accuracy in LoS and NLoS environments for RAR~\cite{zandi2023robot} using CSI.

Another prominent area is using WiFi for robots localization~\cite{zou2020adversarial}. The signal strength ratio (SSR) can be used for simultaneous robot localization and detailed location mapping~\cite{biswas2010wifi}. While GPS and LiDAR have historically been favored in the field of simultaneous localization and mapping (SLAM) for mobile robots, each technology presents its own set of challenges. In indoor environments, GPS utilization often yields significant estimation errors, exacerbated by GPS-denied zones where signal reception is unreliable, rendering it inefficient for real-time localization systems (RTLS). On the other hand, LiDAR, while powerful, faces difficulties in geometrically degraded environments, posing challenges for loop closure and leading to poor performance \cite{ismail2022efficient}.

\section{Proposed Methods} \label{sec: prop method}
% The CSI measurements and images reveals several commonalities in data analysis and machine learning. Despite the differences in their nature and applications, the common thread of data representation, spatial insight, and feature extraction underscores the fundamental approaches to analyzing and learning from CSI measurements. 
In this section, a vision transformer-based model \cite{dosovitskiy2020image}, named bidirectional vision transformer-concatenated (BiVTC), is proposed for RAR using CSI measurements. The vision transformer (ViT) harnesses the potency of self-attention mechanisms, which have demonstrated remarkable success in natural language processing tasks. At its core, the ViT architecture can introduce a novel perspective to the analysis of CSI data. This innovative approach transforms the CSI data into sequences of smaller data segments, similar to patches in image data. This transformation enables harnessing the power of self-attention mechanisms, which excel at capturing intricate relationships between different segments.

Architecture of the proposed BiVTC model is illustrated in Figure \ref{fig:FE-Transforme}.
For each CSI measurement, a patch set is generated by reshaping the input $\mathbf{A}\in \mathbb{R}^{S\times T}$ to $\mathbf{A}_p\in \mathbb{R}^{N\times P \times P}$, where $P$ is height and width of patch $i$, and $N=\frac{S \times T}{P \times P}$ represents the number of patches of a CSI measurement. Then, each patch is flattened and passed to positional encoding and embedding layer which maps the input sequence to a vector of size $L$.

% \begin{figure}[!t]
%     \centering
%     \includegraphics[width=0.45 \textwidth]{floor.png}
%     \caption{Floor plan of the data collection environment. The grid area provides an overview of various sniffer placement options. }
%     \label{fig:floor}
% \end{figure}

The representations of the CSI patches are then channeled through a sequence of transformer layers, each equipped with multi-head self-attention (MHSA), defined as 

\begin{equation}
    \text{MultiHead}(\mathbf{Q}, \mathbf{K}, \mathbf{V}) = \big(\lambda_1\odot \lambda_2 \ldots \odot \lambda_M \big)\mathbf{W}_O,
\end{equation}
where $\odot$ denotes the concatenation operation, $\mathbf{W}_O\in \mathbb{R}^{L\times L}$ is the output weight matrix, and $M$ is the number of attention heads. Each attention head $\lambda_i$ is defined as 
\begin{equation}\label{eq: head}
    \lambda_i = \text{Attention}(\mathbf{Q}\mathbf{W}_{i}^Q, \mathbf{K}\mathbf{W}_{i}^K, \mathbf{V}\mathbf{W}_{i}^V),
\end{equation} 
where $\mathbf{W}_{i}^{Q}, \mathbf{W}_{i}^{K}, \mathbf{W}_{i}^{V}\in \mathbb{R}^{L \times \frac{L}{M}}$ are query, key, and value weight matrices per head, respectively, and 
\begin{equation}
    \text{Attention}(\mathbf{Q}, \mathbf{K}, \mathbf{V}) = \sigma \left(\frac{\mathbf{Q}\mathbf{K}^\top}{\sqrt{d_k}}\right) \mathbf{V},
\end{equation}
where $\top$, $\sigma(\cdot)$, $\mathbf{Q}$, $\mathbf{K}$, and $\mathbf{V}$ denote the transpose operation, Softmax function, query, key, and value matrices, respectively. These values are the amplitude information derived from CSI patches and $d_k$ is the dimension of the key vectors, which captures the essence of relationships between different patches in CSI data. In the final step, the feature vectors $\mathbf{f}_1$ and $\mathbf{f}_2$ are concatenated as
\begin{equation}
        \mathbf{f}_C = \mathbf{f}_1 \odot \mathbf{f}_2,
\end{equation}
where $\mathbf{f}_C$ is passed as an input feature vector to an MLP network with a Softmax layer for dimensionality reduction and classification tasks, respectively. 

The proposed model enables discovering both local patterns and global contexts within the CSI amplitude data, making it proficient at understanding the positions and interactions of key subcarrier components.
The MHSA operation involves concatenating the outputs of attention heads and linearly projecting them to generate the final attention output. It is essential that $L$ and $M$ are selected such that $L$ is divisible by $M$, ensuring $\frac{L}{M}$ yields an integer value. In each transformer block, a multi-layer perceptron (MLP) with Gaussian error linear unit (GeLU) activation function  operates element-wise on each embedded output of MHSA. In the BiVTC model, the computationally efficient rectified error linear unit (ReLU) activation function is utilized. Conversely, the ViT model aggregates these operations across multiple layers to progressively extract hierarchical features from image patches. In this model with two distinct sniffers, two separate ViT networks are utilized, to yield different sets of feature vectors denoted as $\mathbf{f}_1$ and $\mathbf{f}_2$, as illustrated in Figure \ref{fig:FE-Transforme}.

% The mathematical foundation of the ViT model rests on the self-attention mechanism~\cite{dosovitskiy2020image}. Self-attention, represented by the following formula
% \begin{equation}
%     \text{Attention}(\mathbf{Q}, \mathbf{K}, \mathbf{V}) = \sigma \left(\frac{\mathbf{Q}\mathbf{K}^\top}{\sqrt{d_k}}\right) \mathbf{V},
% \end{equation}
% where $\sigma(\cdot)$, $\mathbf{Q}$, $\mathbf{K}$, and $\mathbf{V}$ denote the Softmax function, query, key, and value matrices, which these values are the amplitude information derived from CSI patches, and $d_k$ is the dimension of the key vectors, which captures the essence of relationships between different patches in CSI data.

\section{The RoboFiSense Dataset}
\label{sec: dataset}

\begin{figure}[!t]
    \centering
    \includegraphics[width=0.43 \textwidth]{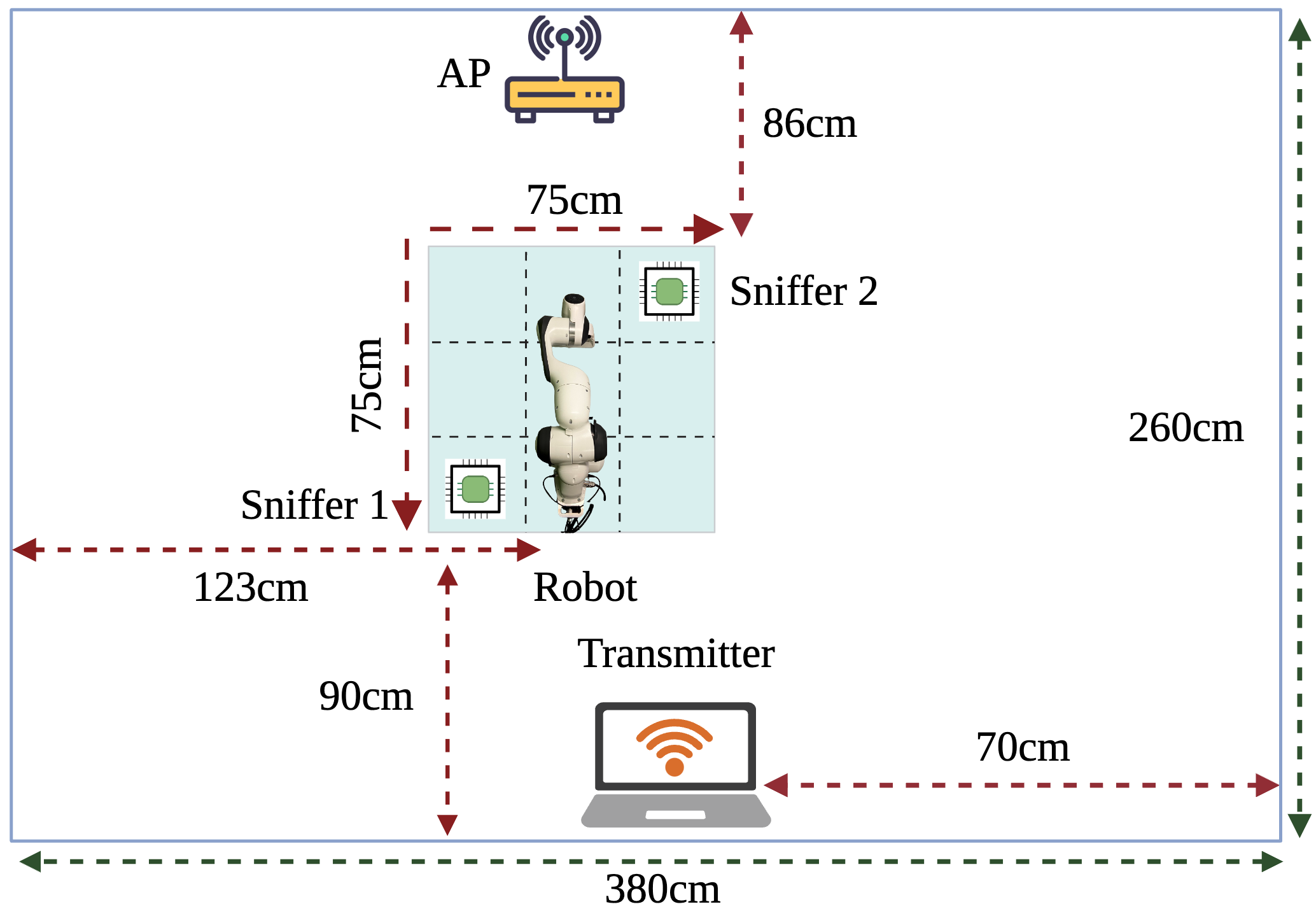}
    \caption{Floor plan of the data collection environment. The grid area provides an overview of various sniffer placement options. }
    \label{fig:floor}
\end{figure}

In this section, the data collection setup and procedure are discussed. Figure~\ref{fig:floor} illustrated the data collection floor plan, where a Franka Emika robotic arm and two sniffers were used to collect the CSI measurement.
\begin{figure}
    \centering
    \subfigure[Hardware setup.]{\includegraphics[width=0.35 \textwidth]{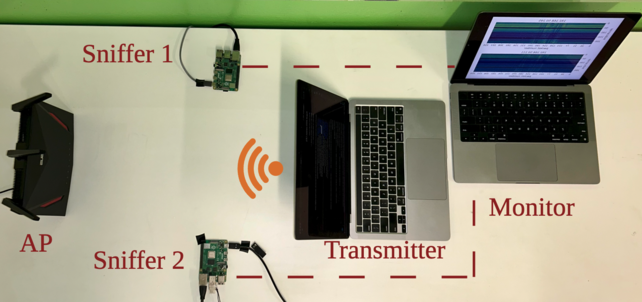}}
    \subfigure[Schematic map of the hardware setup.]{\includegraphics[width=0.35 \textwidth]{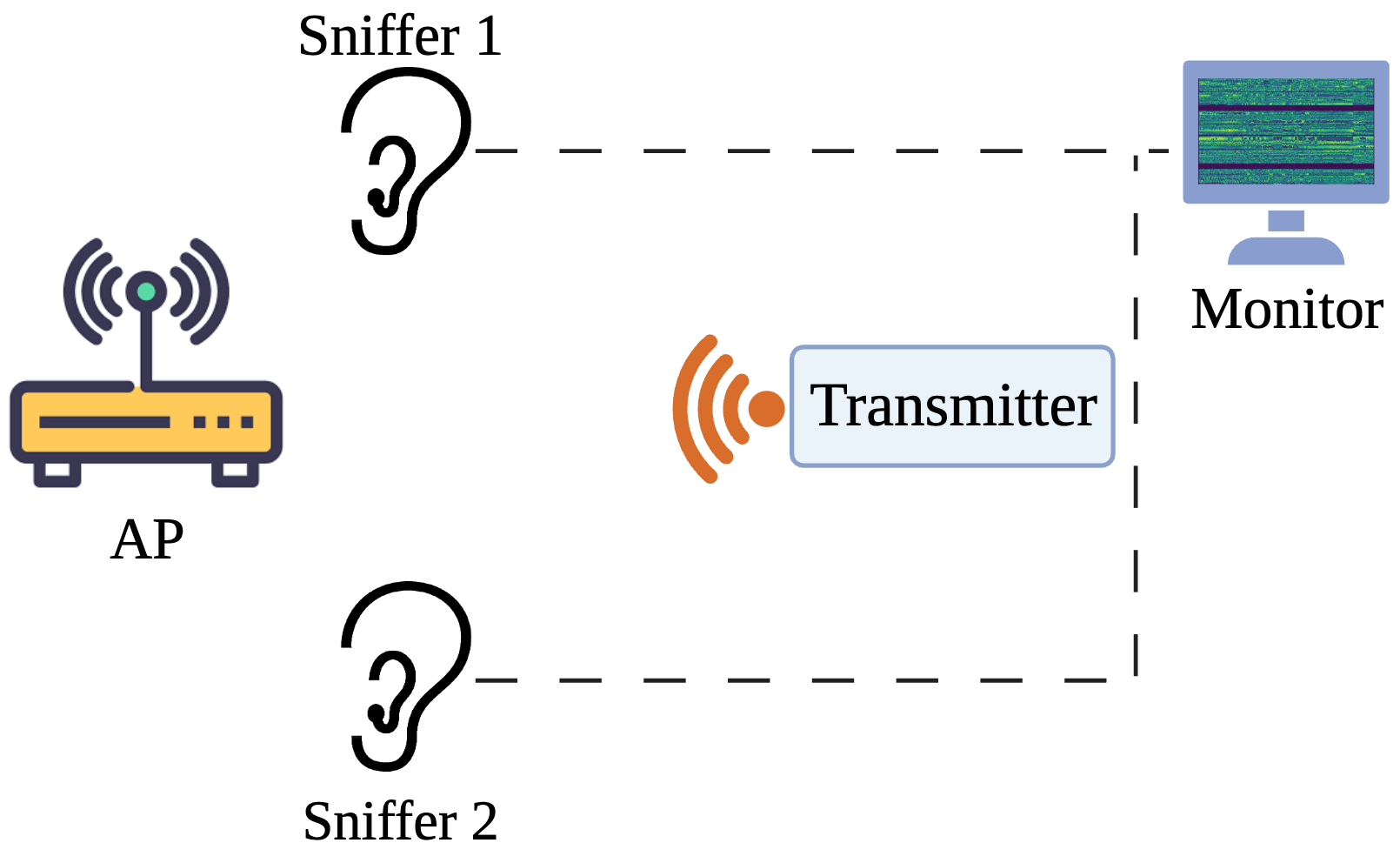}}
    \caption{Illustration of the hardware setup and the corresponding schematic map. }
    \label{fig:communication_chart}
\end{figure}

% The sniffers capture the packets sent by the transmitter. After the CSI extraction is done by the sniffers, the sniffers send the extracted CSI information to the monitor, to be stored, timestamped, and further processed.

\subsection{Hardware Setup}
We employed a dual Raspberry Pi $4$ setup, integrating the Nexmon project \cite{nexmon:project}, to procure comprehensive CSI data from the sniffer devices. Each Raspberry Pi device, equipped with Nexmon, facilitated CSI data acquisition, incorporating local timestamping directly on the hardware. The timestamped data is then sent to the loop-back interface of the network.  While this configuration effectively captures data from individual sniffers, the synchronization of timestamps becomes crucial for multi-sniffer deployments. Addressing this requirement, we devised a solution where loop-back packets containing CSI data from each Raspberry Pi are redirected to a dedicated system, referred to as the monitor.

The monitor functions as a central hub, gathering packets from various sniffers and applying synchronized timestamps as dictated by a designated frequency. A visual representation of this intercommunication is presented in Figure~\ref{fig:communication_chart}. It is worth noting that in line with the Nexmon project's specifications, Raspberry Pis operating as sniffers lose WiFi communication capability. To circumvent this limitation, we interconnect the sniffers and the monitor via Ethernet cables, ensuring seamless communication between the sniffers and the monitor.

\begin{figure}[!t]
\centering
\subfigure[Arc]{\includegraphics[width=0.2 \textwidth]{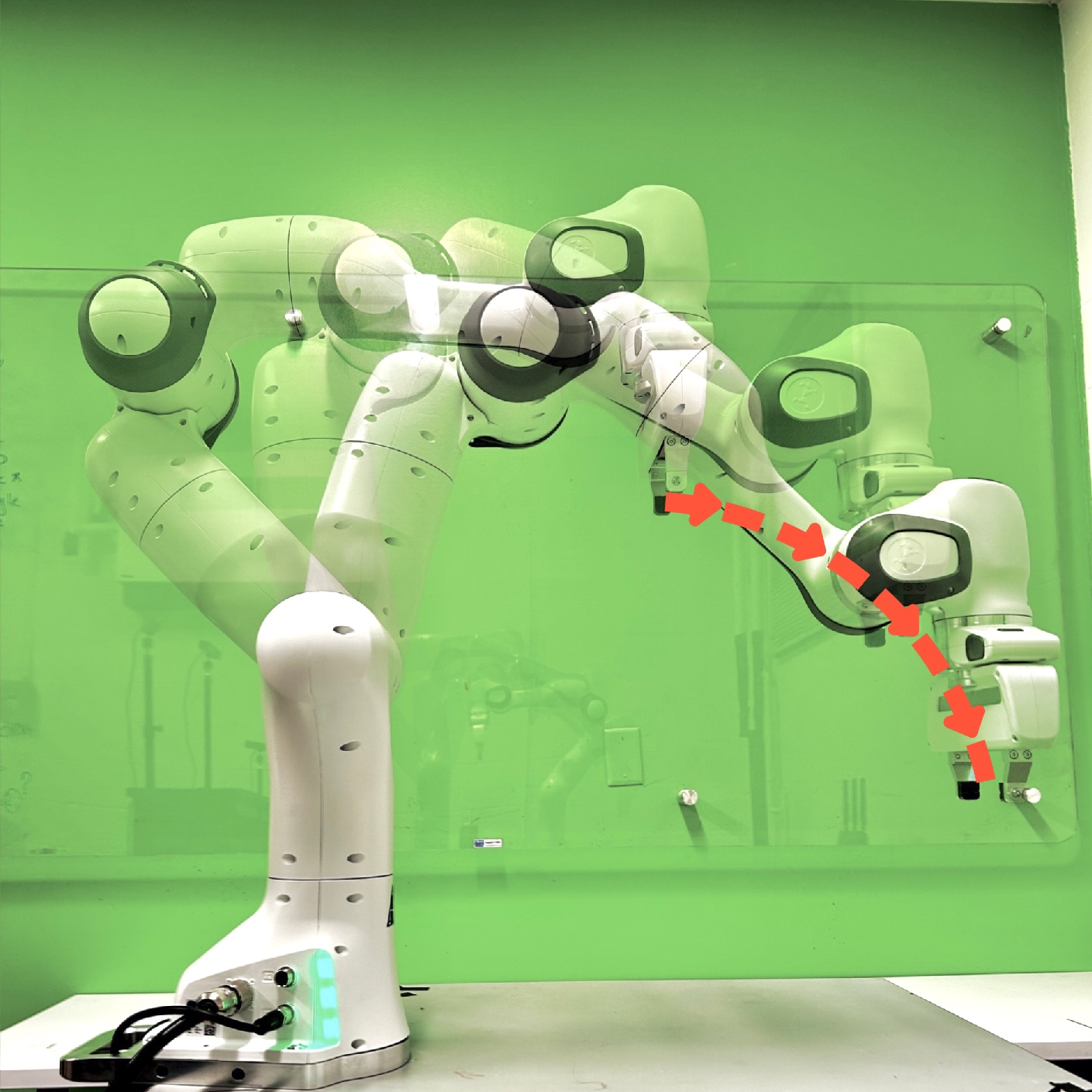}}
\subfigure[Elbow]{\includegraphics[width=0.2 \textwidth]{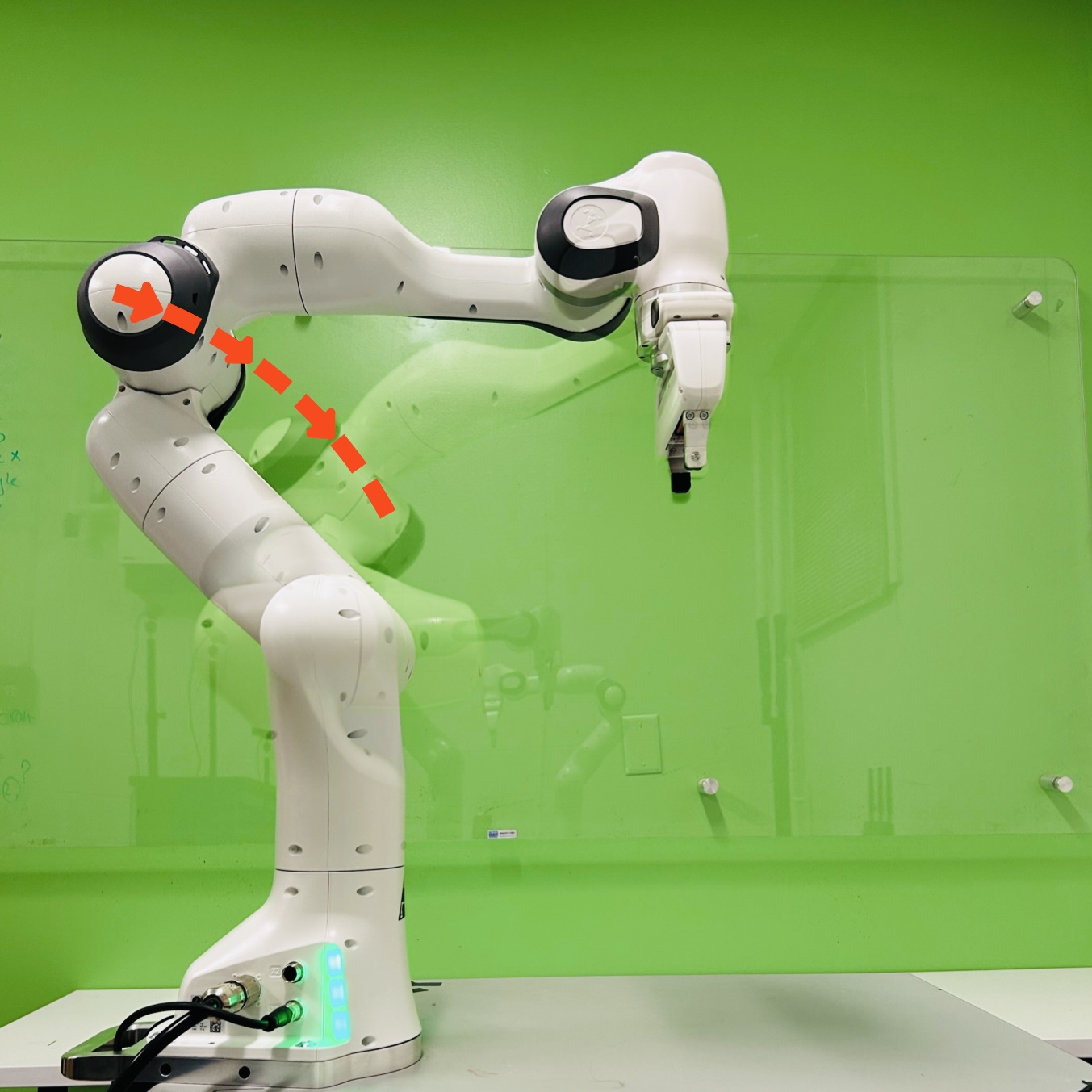}}

\subfigure[Rectangle]{\includegraphics[width=0.2 \textwidth]{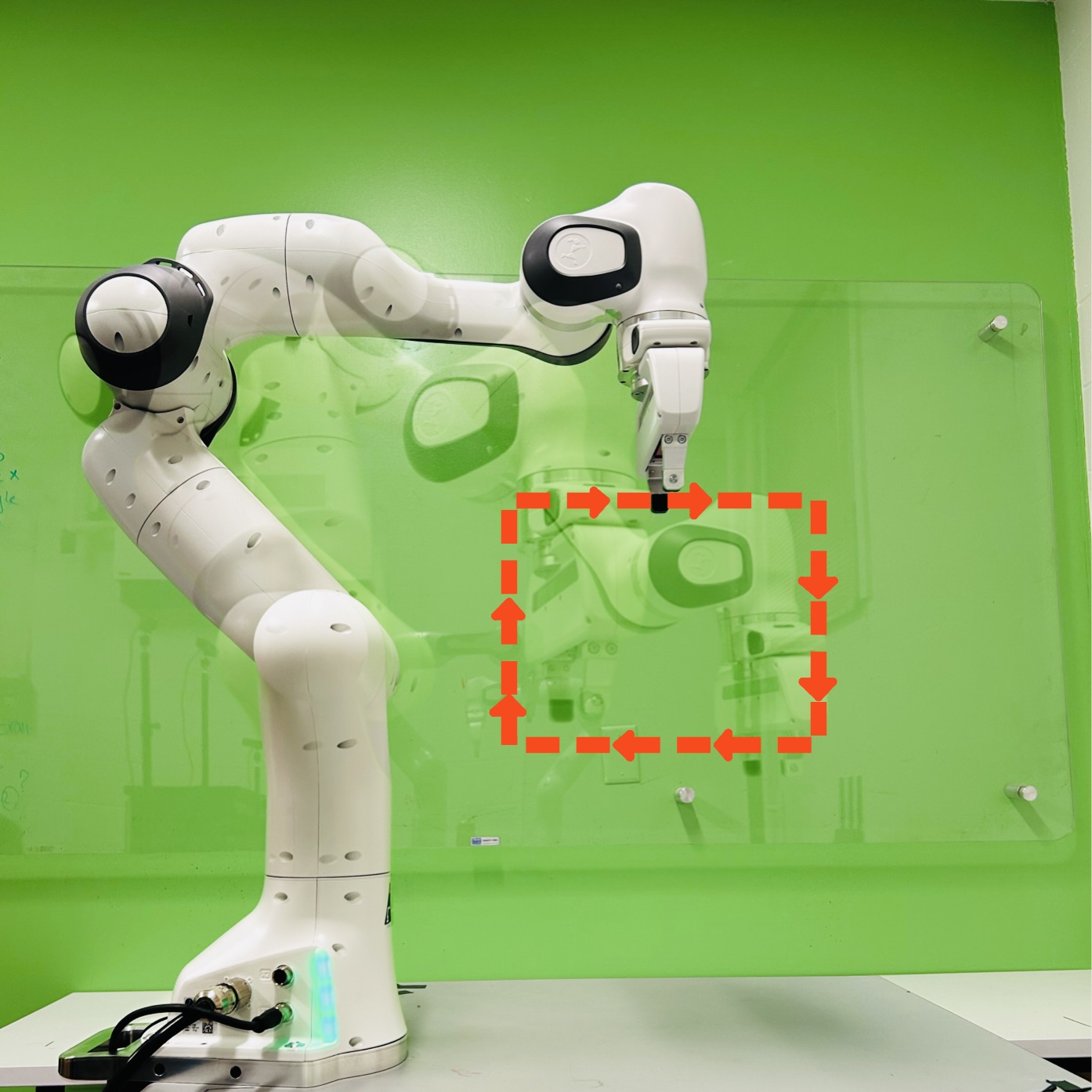}}
\subfigure[Silence]{\includegraphics[width=0.2 \textwidth]{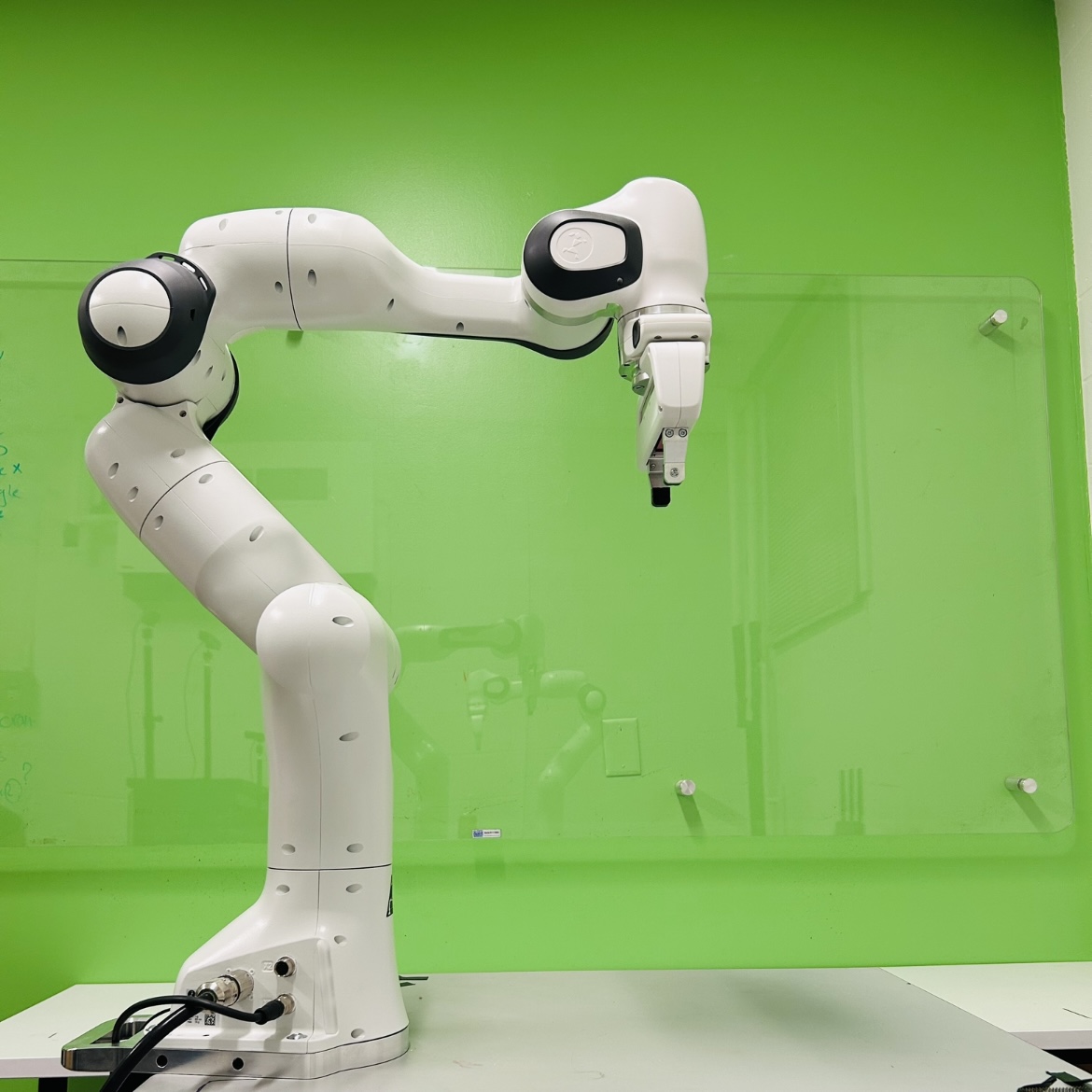}}

\subfigure[SL-Forward]{\includegraphics[width=0.2 \textwidth]{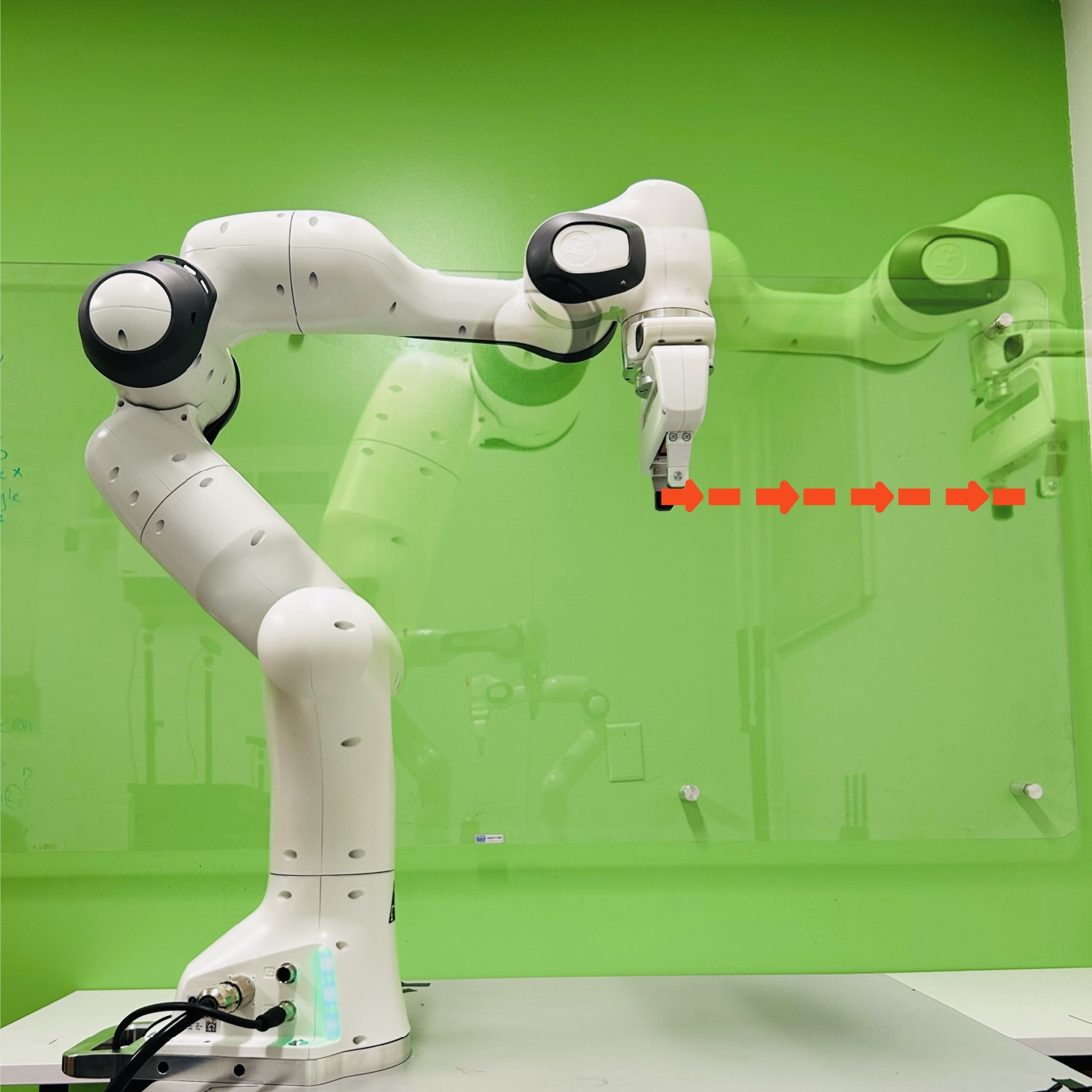}}
\subfigure[SL-Right Left]{\includegraphics[width=0.2 \textwidth]{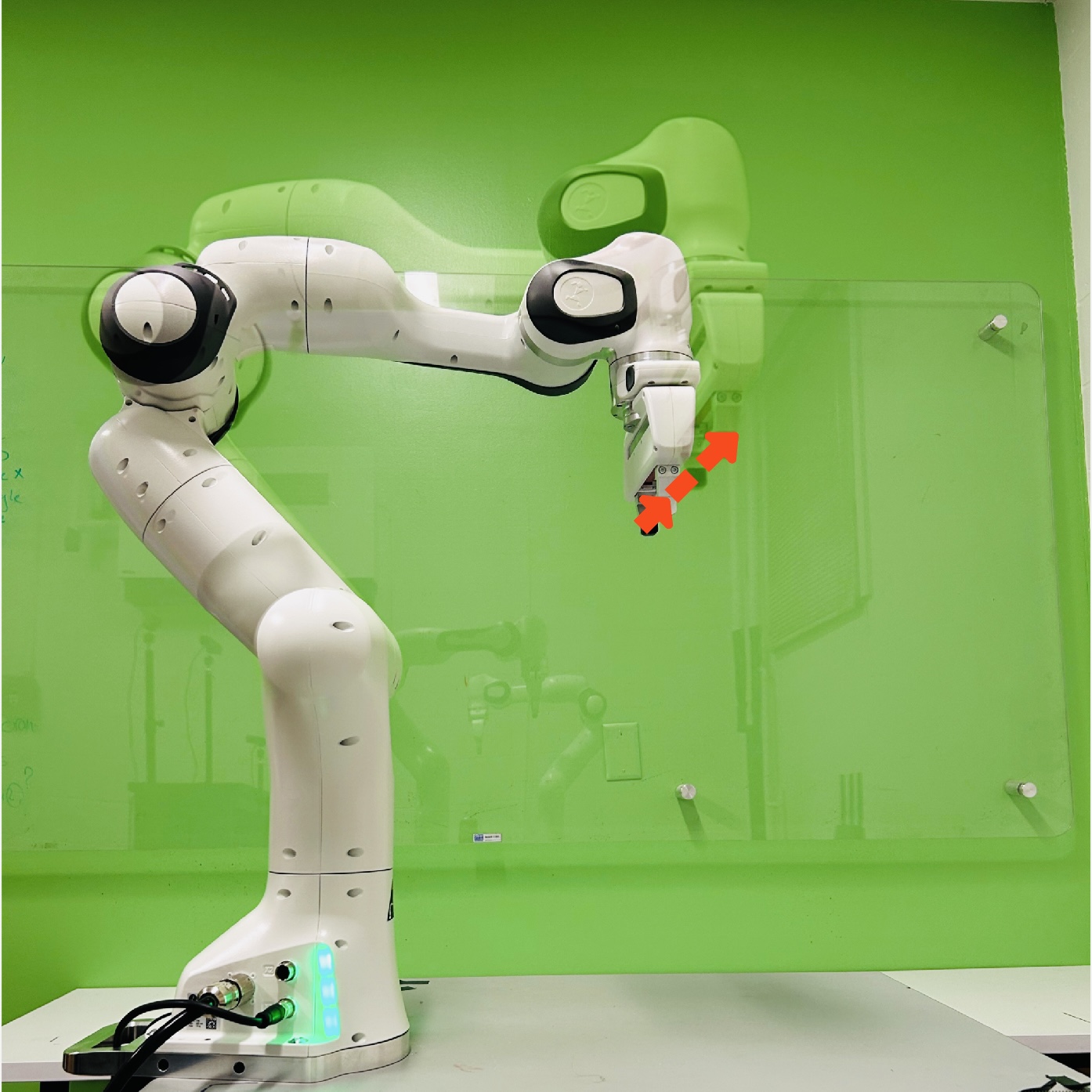}}

\subfigure[SL-Up Down]{\includegraphics[width=0.2 \textwidth]{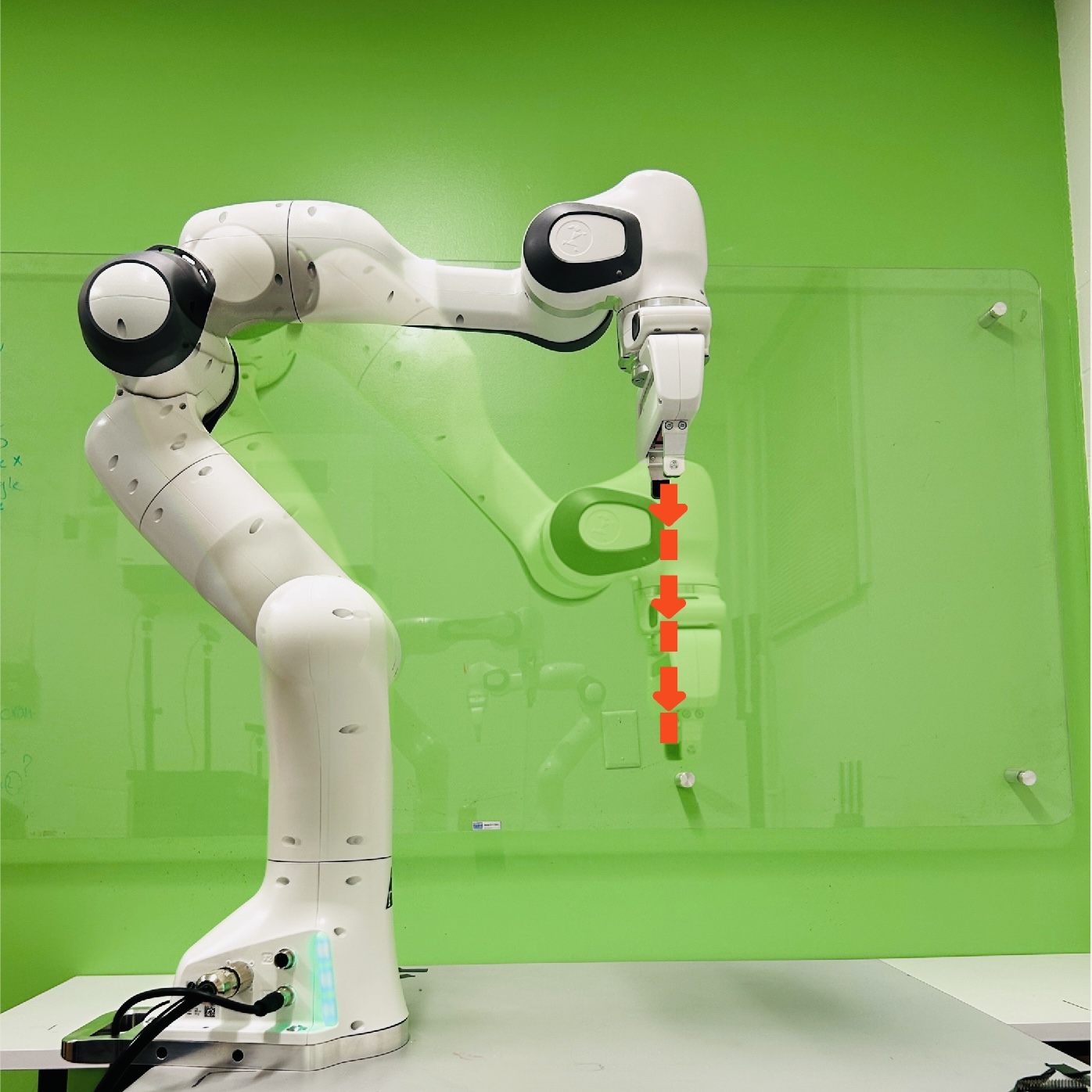}}
\subfigure[Triangle]{\includegraphics[width=0.2 \textwidth]{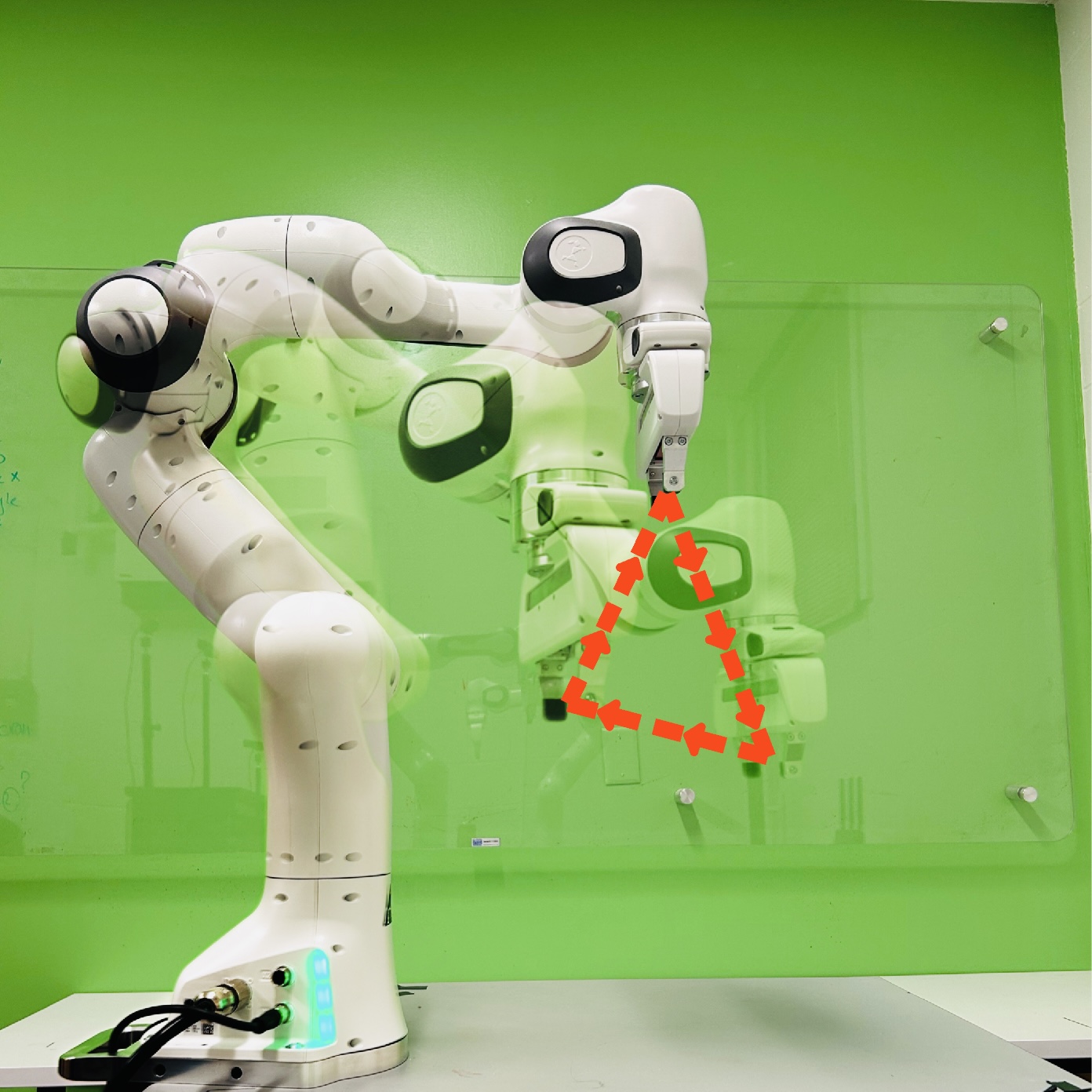}}
\caption{Illustration of the eight activities performed by the Franka Emika arm in the experiments: (a) Arc, (b) Elbow, (c) Rectangle, (d) Silence, (e) Straight Line - Forward (SLFW), (f) Straight Line - Right Left (SLRL), (g) Straight Line - Up Down (SLUD), and (h) Triangle. The motion patterns of the robotic arm are shown with red dashed lines.}
\label{fig:actions}
\end{figure}

\begin{figure}[!t]
\centering
\subfigure[Arc activity.]{\includegraphics[width=0.4 \textwidth]{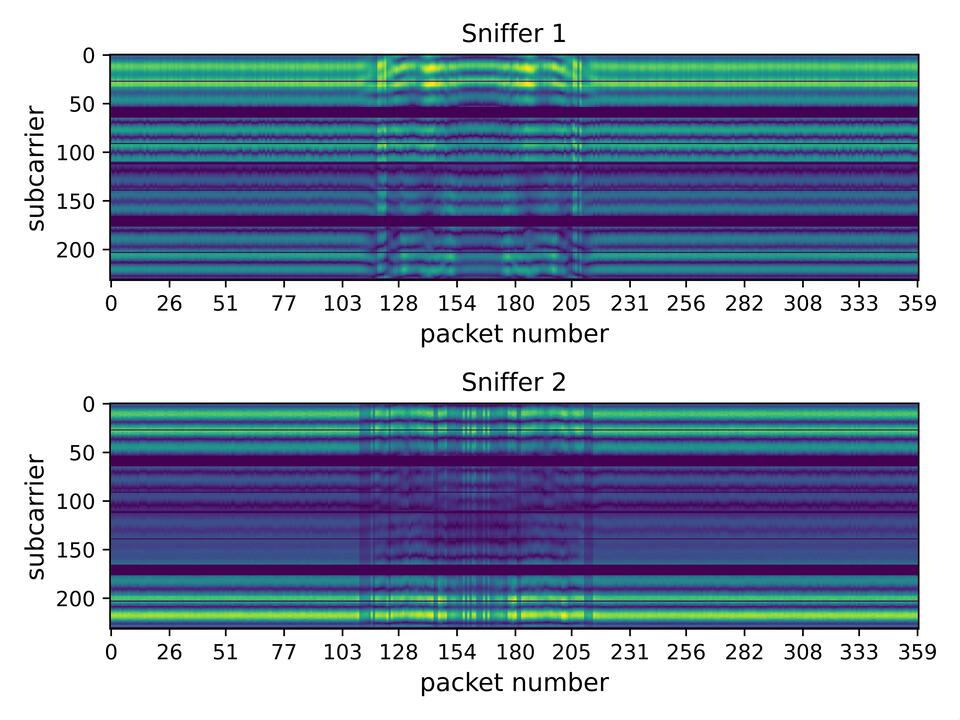}}

\subfigure[Rectangle activity.]{\includegraphics[width=0.4 \textwidth]{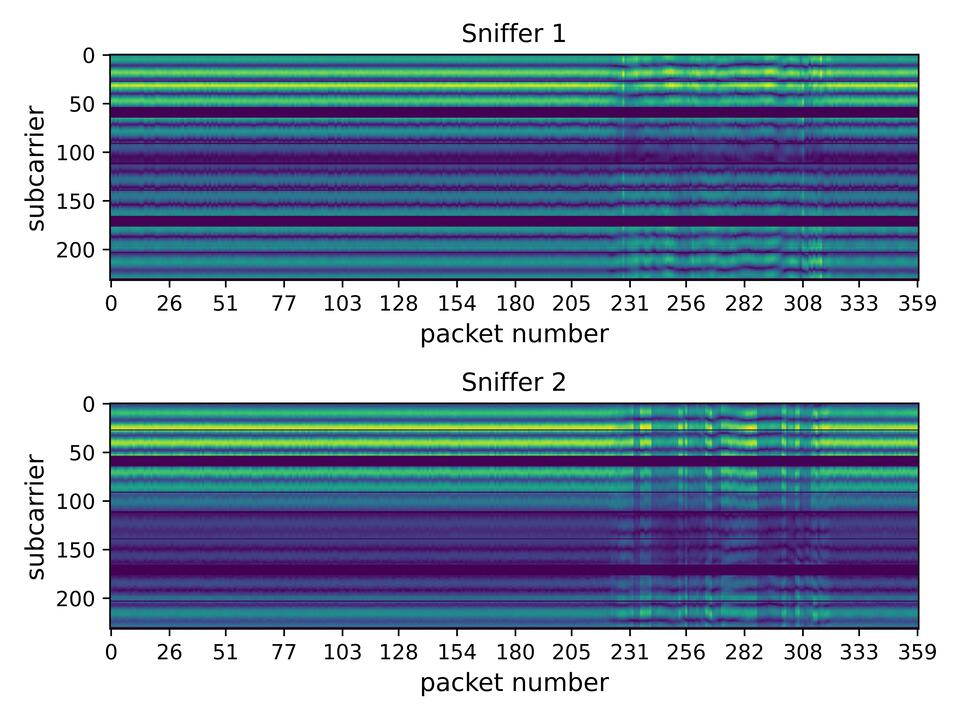}}
\caption{Synchronized CSI collection from two sniffers during robot's "Arc" and "Rectangle" activities.}
\label{fig:synchronized_time}
\end{figure}

\subsection{Data Collection Procedure} \label{sec: data col}

\subsubsection{Activity Classes}
The Franka Emika robot arm was programmed to perform eight different activities of a Franka Emika robot; (a) Arc, (b) Elbow, (c) Rectangle, (d) Silence, (e) Straight Line - Forward (SLFW), (f) Straight Line - Right Left (SLRL), (g) Straight Line - Up Down (SLUD), and (h) Triangle. The activity paths of the robotic arm are shown with red dashed lines in Figure~\ref{fig:actions}. 

The color-maps in Figure~\ref{fig:synchronized_time} show the CSI amplitude values extracted from sniffers $1$ and $2$ for the Arc and Rectangle activity classes. This visual representation also showcases the synchronization between sniffers, capturing instances of CSI packet distortions resulting from sudden activities during a specific time interval, as observed by both CSI sniffers.

\subsubsection{CSI Measurements}
The process of data collection encompassed the acquisition of CSI through two sniffers placed at different locations within the room.
%The $80$ MHz bandwidth and a sampling rate of $30$ Hz were used in each sniffer to capture $256$ sub-carriers at every timestamp~\cite{10.5555/2563615,  meneghello2023csi}. 
The 802.11ac standard offers the choice of $20$ MHz, $40$ MHz, and $80$ MHz bandwidths \cite{10.5555/2563615}, and these
options correspond to $64$, $128$, and $256$ subcarriers, respectively. To ensure the capture of the maximum number of subcarriers at each timestamp, we have selected $80$ MHz bandwidth option \cite{meneghello2023csi}. This decision enabled each sniffer to capture the maximum number of subcarriers at each timestamp, facilitating a more detailed acquisition of CSI crucial for RAR.
 The CSI measurements were collected over a $12$-second interval, yielding a collection of CSI matrices denoted as $\mathbf{H}\in \mathbb{C}^{360\times 256}$ prior to preprocessing. The preprocessing steps include exclusion of pilot and unused subcarriers \cite{10.5555/2563615} followed by  computation of CSI amplitudes. Consequently, this process led to a reduction in matrix size, yielding $\mathcal{\mathbf{A}}\in \mathbb{R}^{360\times 236}$ for each sample. The most prolonged activity sequence in our dataset spanned up to four seconds, with the occurrence of activities transpiring at random intervals within the $0$ to $8$-second time-frame.

% This bandwidth choice enables using a larger number of subcarriers and is strategic, aiming to enhance the granularity of the data collected while maintaining noise at manageable levels, thereby optimizing our system's ability to accurately recognize complex activities \cite{9367546}. 

\subsubsection{Robotic Arm Velocity}
\label{sec:velocity_data}
As it is discussed in~\cite{hasanzadeh2023hand}, hand velocity has a direct impact on the performance of machine learning models for HAR. Hence, to study the impact of robotic arm velocity while performing an activity, the CSI measurements were collected based on three distinct velocity levels, each of which incorporates a $10\%$ increase in both velocity and acceleration. 
The dataset for each activity was collected at three different velocities: $v_1$, $v_2$, and $v_3$, with the datasets corresponding to each velocity denoted as $\mathcal{V}_1$, $\mathcal{V}_2$, and $\mathcal{V}_3$, respectively. Here, $v_1$ represents the slowest pace, $v_2$ is $10\%$ faster, and $v_3$ increases by $20\%$ relative to $v_1$, so $v_1 < v_2 < v_3$.

\subsubsection{Sniffers Placement}
In order to study sensitivity of the machine learning models to sniffer location, the CSI measurements were collected from different combination of sniffers placements. To do so, a grid consisting of nine unique locations was used as demonstrated in Figure \ref{fig:floor}. In each experiment, the two sniffers were strategically relocated to different positions within this grid. Notably, one location within the grid was continually occupied by our stationary robot, ensuring that it remained static. This configuration allowed us to explore four distinct scenarios of data collection.

\begin{table}[!t]
\footnotesize
    \centering
    \caption{Classification performance (Average$\pm$Std) of the ConvLSTM, CNN, LSTM, BiLSTM, ViT and BiVTC models after 5-fold cross-validation, in percentage (\%).}
    %\vspace{0.1cm}
    \begin{tabular}{|c|c|c|c|c|}
    \hline
    \scriptsize{Model} & Precision  & Recall& F1-Score & Accuracy\\
    \hline \scriptsize{ConvLSTM} &77.74$\pm$2.33 & 77.50$\pm$2.85 & 76.25$\pm$2.59 & 77.50$\pm$2.85 \\
    \hline \scriptsize{CNN} & 84.63$\pm$3.19 & 83.15$\pm$2.36 & 82.43$\pm$3.47 & 83.15$\pm$2.36\\
    \hline \scriptsize{LSTM} & 89.43$\pm$1.89 & 88.75$\pm$2.03 & 88.32$\pm$1.94 & 88.75$\pm$2.03\\
     \hline \scriptsize{BiLSTM} & 90.49$\pm$1.26 & 89.58$\pm$1.78 & 87.17$\pm$1.48 & 89.58$\pm$1.78\\
    \hline \scriptsize{ViT} & 91.27$\pm$1.22 & 90.76$\pm$2.61 & 90.51$\pm$2.78 & 90.76$\pm$2.61\\
    \hline \scriptsize{BiVTC} & \bf{93.33}$\pm$2.23 & \bf{92.50}$\pm$2.45 & \bf{92.45}$\pm$2.91 & \bf{92.50}$\pm$2.45\\
    \hline
    \end{tabular}
    \label{tab:sep-sc}
\end{table}

\section{Experiments} % \& Results Analysis}
\label{sec:exp}
In this section, the proposed BiVTC model and other state-of-the-art machine learning models are evaluated for RAR in various scenarios.

\subsection{Machine Learning Models} 
\label{sec: Model div}
The machine learning models used for RAR in the experiments are CNN \cite{gu2018recent}, convolutional long short-term memory (ConvLSTM) \cite{shi2015convolutional}, multi-variate long short-term memory (LSTM) \cite{karim2019multivariate}, bidirectional LSTM (BiLSTM) \cite{graves2005bidirectional}, ViT \cite{dosovitskiy2020image}, and the proposed BiVTC model. 
The CNN model has a series of convolutional, max-pooling, and fully connected layers, along with appropriate regularization techniques~\cite{zandi2023robot}.

% In the architecture of the CNN model, the initial layers consist of convolutional layers that employ ReLU activation functions to introduce non-linearity \cite{agarap2018deep}. The subsequent layers are equipped with max-pooling operations to down-sample the spatial dimensions, aiding in reducing the computational complexity while retaining the most relevant features.
% The architecture culminates in a fully connected layer that flattens the extracted features and feeds them into a dense layer with ReLU activation. This layer aims to aggregate and consolidate the learned representations before leading to the final classification layer. The classification layer, consisting of a softmax activation function, assigns class probabilities to each image sample.

In the CNN model, the ${L}_2$ and dropout regularization techniques are integrated. The regularization coefficients are empirically determined.  
The multivariate LSTM model has two stacked LSTM layers, and each layer has $64$ features in its hidden state. Similar hyperparameters and number of layers are also used for the BiLSTM model, after performing a grid-search. The ConvLSTM model employs a single-layer architecture with a hidden dimension of 64, representing the number of features within the ConvLSTM cells. The convolution operations utilize a $(3, 3)$ kernel size. The model incorporates global average pooling followed by a fully connected layer.

The patch size for the ViT model is set to $45$, determining the dimensions of CSI patches used for processing. A batch size of $16$ is employed during training to balance computational efficiency and model convergence.  
The learning rate of $1\times10^{-4}$ is used with the weight decay set to $2\times10^{-5}$. The ViT architecture incorporates four attention heads and a stack of six transformer layers to capture intricate spatial dependencies within the CSI data. Dropout regularization with a rate of $0.4$ is applied throughout the network to enhance generalization and prevent overfitting. These hyperparameters are fine-tuned to ensure the ViT model's optimal performance on CSI data classification tasks. The BiVTC model has two vision transformers, each tailored to capture distinctive features from the different sniffers. These features are then fused through concatenation, providing the model with a richer representation to enhance classification accuracy. This capacity is pivotal in recognizing unique spatial structures present in the data.

\subsection{Training and Evaluation Setup}
All models in each experiment underwent 5-fold cross-validation. The reported results include the average performance and standard deviation of accuracy, precision, recall, and F1-score, where applicable. Within each cross-validation fold, the dataset underwent shuffling and was divided into training $70\%$, validation $10\%$, and testing $20\%$ subsets. Hyperparameters for each model were selected by grid search using the validation dataset. An early stopping mechanism with a patience of $15$ epochs was used to mitigate overfitting during $150$ training epochs for all the models. Performance of each model is evaluated using categorical cross-entropy loss and accuracy metrics during both the training and validation phases. 

\begin{figure*}[!t]
    \centering
\subfigure[Trained on $\mathcal{V}_1 \& \mathcal{V}_2$ - Tested on $\mathcal{V}_3$]{\includegraphics[width=0.32 \textwidth]{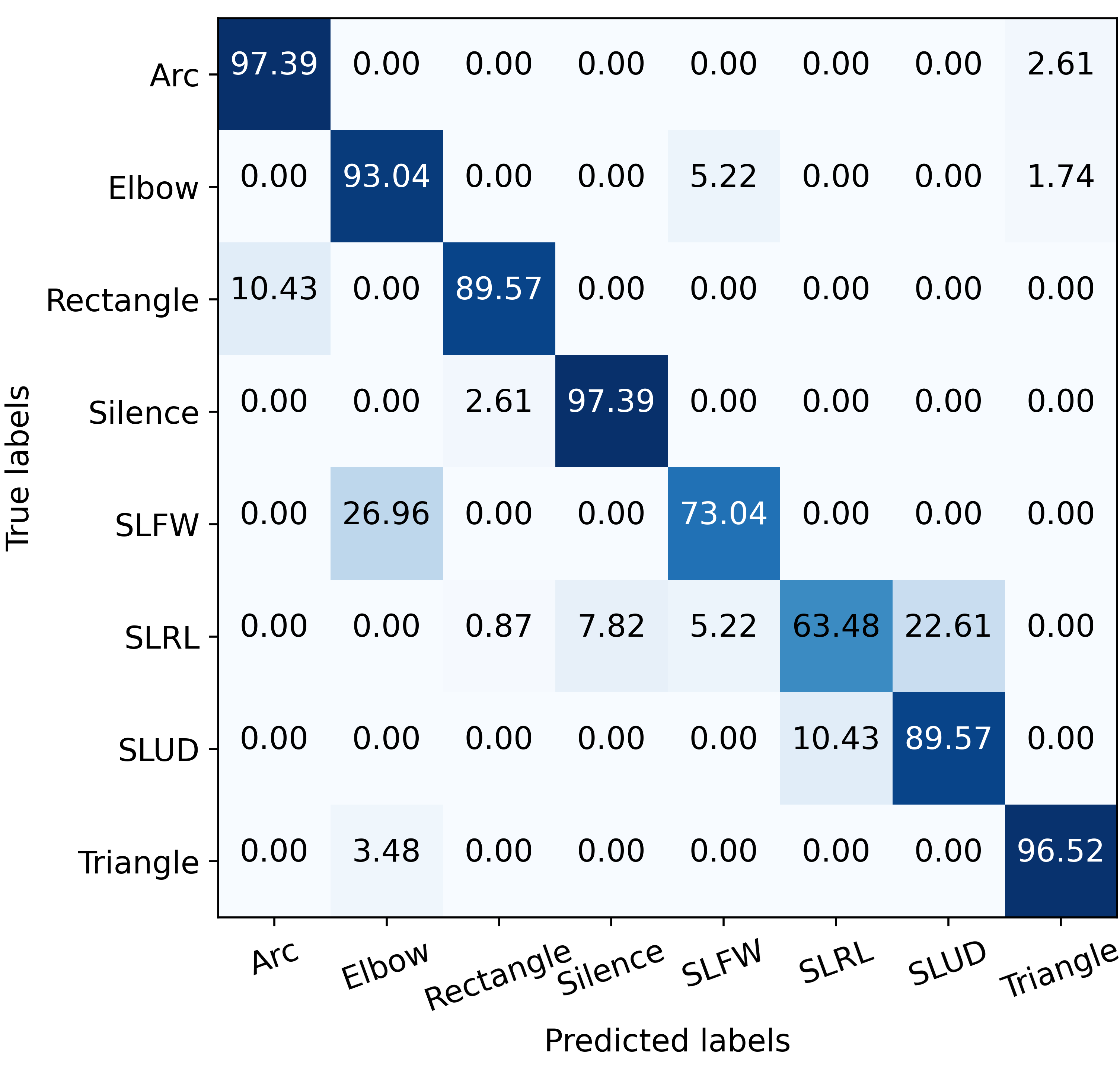}}
\subfigure[Trained on $\mathcal{V}_1 \& \mathcal{V}_3$ - Tested on $\mathcal{V}_2$]{\includegraphics[width=0.32 \textwidth]{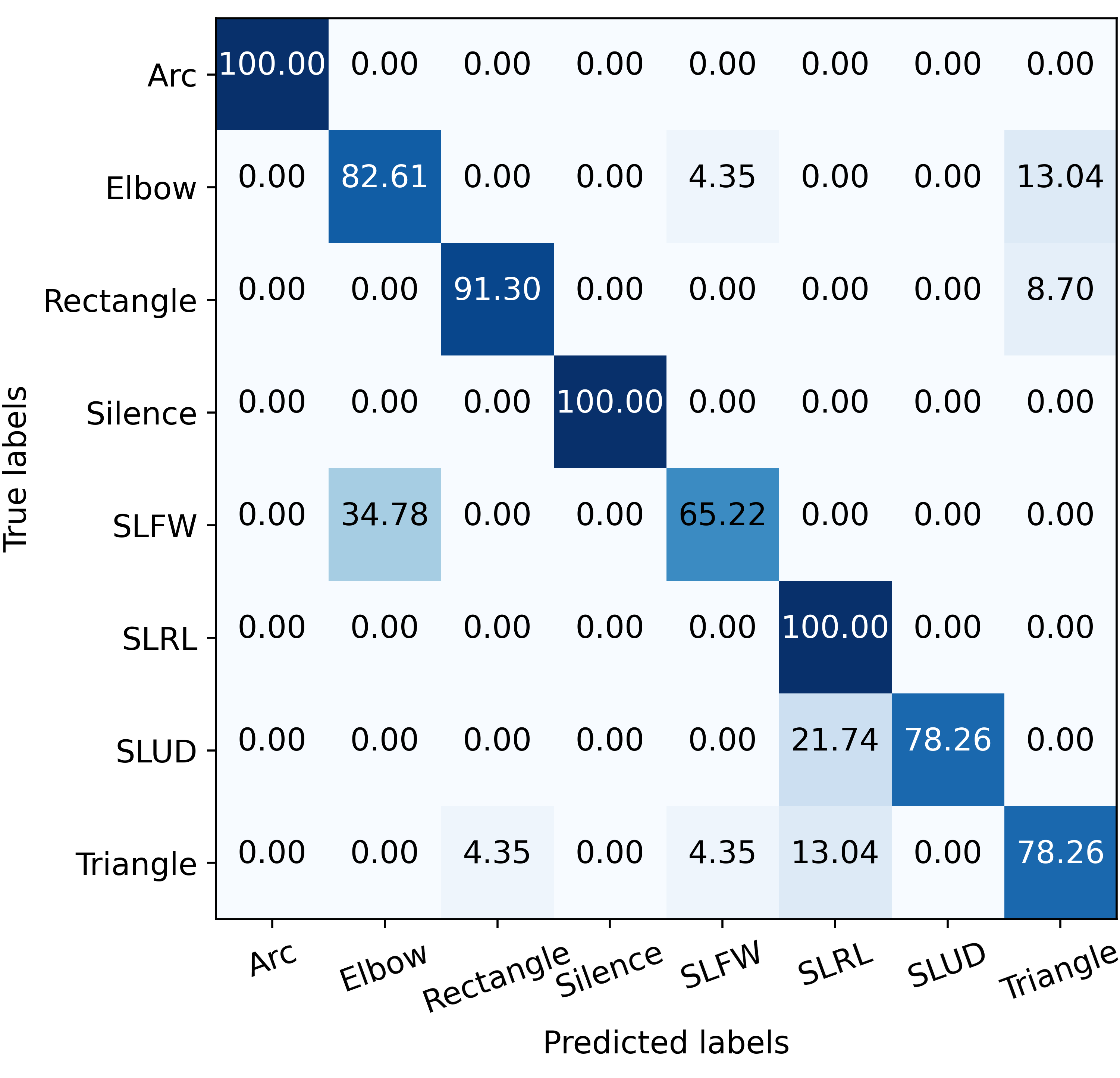}}
\subfigure[Trained on $\mathcal{V}_2 \& \mathcal{V}_3$ - Tested on $\mathcal{V}_1$]{\includegraphics[width=0.32 \textwidth]{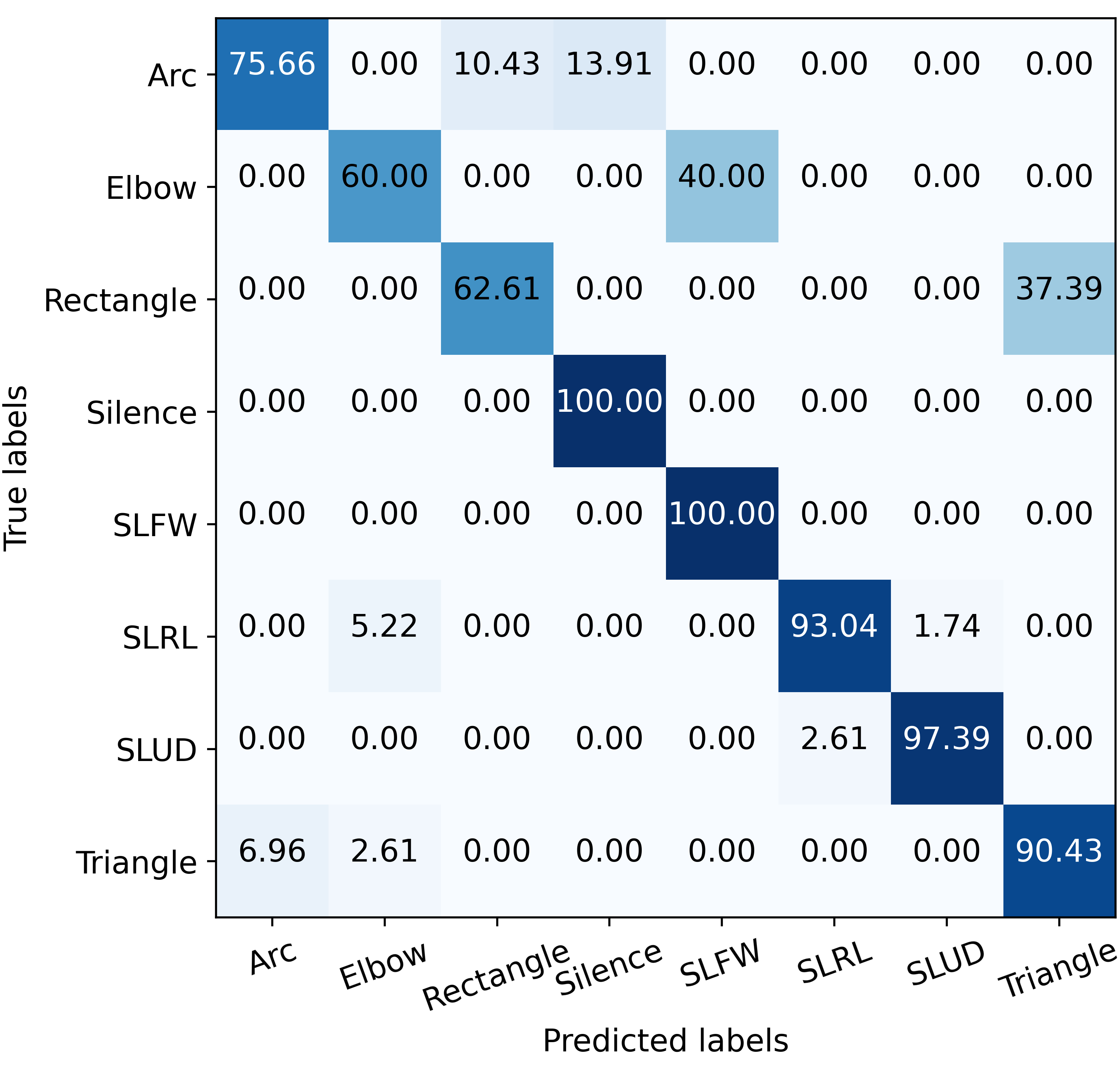}}
\caption{Confusion matrices of the proposed BiVTC model for the leave-one-velocity-out cross-validation experiments.}
\label{fig: conf V}
\end{figure*}

\begin{table*}[!t]
\centering
\caption{Classification performance of the models for the leave-one-velocity-out cross-validation study per activity class in percentage (\%).}
\begin{tabular}{|c|c||c|c|c|c|c|c|c|c|c|c|}
\hline Train & Test & Model & Arc & Elbow & Rectangle & Silence & SLFW & SLRL& SLUD& Triangle & All\\
\hline \multirow{3}{*}{$\mathcal{V}_1$ \& $\mathcal{V}_2$}&  \multirow{3}{*}{$\mathcal{V}_3$}& CNN & 86.21 & 26.09 & 56.52 & 83.02 & 28.62 & 34.78 & 86.96 & 84.92 & 60.33 \\
&  & ViT & 78.26 & 89.14 & 73.26 & 78.26 & 50.00 & 63.04 & 30.43 & 82.61 & 68.20\\
& & LSTM & 91.30 & 60.87 & 56.52 & 95.65 & 97.10 &63.76 & 66.67 & 60.87& 74.09\\
& & BiLSTM & 95.65 & 44.43 & 35.73 & 94.7 & 91.30 & 94.7 & 82.61 & 86.96 & 78.26\\
 &  & BiVTC & 97.39 & 93.04 & 89.57 & 97.39 & 73.04 & 63.48 & 89.57 & 96.52 & \textbf{87.50} \\
\hline
\hline \multirow{3}{*}{$\mathcal{V}_1$ \& $\mathcal{V}_3$}&\multirow{3}{*}{$\mathcal{V}_2$} & CNN & 65.22 & 82.61 & 39.13 & 100.00 & 100.00 & 17.39 & 43.48 & 95.65 &  67.93\\
 & & ViT & 86.96 & 47.83 & 52.17 & 91.30 & 100.00 & 82.61 & 91.30 & 26.09 & 72.28\\
 & & LSTM &71.73 & 54.35 & 54.35 & 89.13 & 78.26 & 84.78 & 86.95 & 84.78 & 74.41 \\
& & BiLSTM & 96.30 & 69.57 & 91.30 & 96.30 & 95.65 & 100.00 & 51.62 & 51.62 & 81.52\\
 &  & BiVTC & 100.00 & 82.61 & 91.30 & 100.00 & 65.22 & 100.00 & 78.26 & 78.26 & \textbf{86.96} \\
\hline
\hline \multirow{3}{*}{$\mathcal{V}_2$ \& $\mathcal{V}_3$}&\multirow{3}{*}{$\mathcal{V}_1$}  & CNN & 67.80 & 21.74 & 30.43 & 100.00  & 73.91 & 73.91 & 86.96 & 73.91 & 68.47 \\
 &  & ViT & 39.13 & 73.91 & 43.48 & 95.65 & 100.00 & 52.17 & 95.65 & 47.83 & 68.48\\
 & & LSTM & 75.36 & 100.00 & 63.77 & 89.86 & 62.32 & 50.72 & 65.22 & 84.06 & 73.91 \\
& & BiLSTM & 100.00 & 95.65 & 91.30 & 86.96 & 52.17 & 34.78 & 86.96 & 100.00 & 80.98\\
 &  & BiVTC & 75.65 & 60.00 & 62.61 & 100.00 & 100.00 & 93.04 & 97.39 & 90.43 & \textbf{84.89}\\
\hline
\end{tabular}
\label{tab:vel classes}
\end{table*}

\subsection{Baseline Classification Performance}
Table \ref{tab:sep-sc} presents the performance of machine learning models in classification of robotic arm activities. The standout finding is the consistent superiority of the proposed BiVTC model, with an accuracy of $92.50\%$ and F1-Score of $92.45\%$ over the others. Its dual-stream architecture processes inputs separately before integration, enhancing its ability to discern variations across spatial locations. This approach is crucial, especially in scenarios with distinct periods of silence and activity in CSI data. Attention-based models like BiVTC and ViT excel here by selectively focusing on relevant data segments, boosting detection and classification accuracy. BiVTC preserves each sniffer's data integrity, facilitating more precise feature learning and enhanced classification. While LSTM-based models excel in handling time-series data, ViT and BiVTC's complexity and parameter tuning offer an edge in scenarios requiring sensitivity to spatial-temporal dynamics. Their superior F1-scores and accuracy underscore the importance of aligning model architecture with task specifics.

\subsection{Performance Evaluation for Different Arm Velocities} 
\label{dif vel}
Hand movement velocity plays an important role in HAR~\cite{hasanzadeh2023hand}. In order to evaluate the understudy models for RAR with various robotic arm velocities, we collected CSI measurements for various arm velocities performing the eight activities and trained and evaluated the models. This approach allows for meticulous examination of the dataset under carefully controlled velocity conditions.
In this series of experiments, we focused on maintaining consistent locations and activity classes, allowing variations only in speed and acceleration parameters to test the models' robustness. Each model was trained and evaluated based on a leave-one-velocity-out (LOVO) cross-validation strategy, which refers to training a model on two velocity subsets and validation on a third subset, as discussed in Section~\ref{sec:velocity_data}. 

Table~\ref{tab:vel classes} shows the accuracy of machine learning models for different combinations of training and test datasets per activity class. The performance results show that the proposed method has a more generalization performance against unseen arm velocities compared to the other models in a LOVO cross-validation scheme. For example, the proposed BiVTC model has an overall accuracy of $87.50\%$ compared to the BiLSTM model of accuracy $78.26\%$ when trained on the $\mathcal{V}_1$ and $\mathcal{V}_2$ datasets and tested on the $\mathcal{V}_3$ dataset.
The confusion matrix for each LOVO cross-validation experiment for the best (i.e. the proposed BiVTC) model presented in Figure~\ref{fig: conf V}. Based on these matrices, a more detailed evaluation of the proposed BiVTC model with respect to the precision, recall, and F1-score metrics is presented in Table~\ref{tab:vel}.

% Based on the results in subsection \ref{sec: Model div}, we employed BiVTC, which had the best performance in comparison with the other models. Notably, BiVTC demonstrates promising results in the recognition of robotic activity under diverse velocity and acceleration conditions. To have a better understanding of the BiVTC model performance we present classification metrics' of the model in Table \ref{tab:vel} and the confusion matrices of the model tested on three different velocities, in Figure \ref{fig: conf V} (see the Appendix).

% To have a fair comparison between the models, we applied LOVO cross-validation on our top five models, based on results of Table \ref{tab:sep-sc}. Table \ref{tab:vel classes}, presents the accuracy of each class derived from LOVO cross-validation across the top five models, reflecting how each model adapts to different speed conditions. As can be seen, our model, BiVTC, demonstrates superior robustness against velocity changes, particularly in the RAR scenario.

\begin{table}[!t]
\footnotesize
    \centering
    \caption{Detailed performance of the best (i.e. the proposed BiVTC) model  trained and tested based on the leave-one-velocity-out cross-validation scheme in percentage (\%).}
    %\vspace{0.2cm}
    \begin{tabular}{|c|c||c|c|c|c|}
    \hline
    Train & Test &Precision & Recall & F1-Score & Accuracy \\
    \hline $\mathcal{V}_1$ \& $\mathcal{V}_2$ & $\mathcal{V}_3$ & 88.71& 87.50 & 87.08 & 87.50\\
    \hline $\mathcal{V}_1$ \& $\mathcal{V}_3$ & $\mathcal{V}_2$ &  88.31& 86.96 & 86.95 & 86.96\\
    \hline $\mathcal{V}_2$ \& $\mathcal{V}_3$ & $\mathcal{V}_1$ & 86.46 & 84.89 & 84.42 & 84.94\\
    \hline
    \end{tabular}
    \label{tab:vel}
\end{table}

\subsection{Sampling Frequency Analysis} \label{dif freq}
 Sampling frequency of CSI measurements directly influences the level of details and temporal resolution of the captured signals, thereby affecting the system's ability to detect subtle variations in robotic movements. To address this, our study meticulously examines the impact of various sampling rates on the quality and efficacy of the recognition process. 

The benchmark dataset was initially collected at a baseline frequency of $30$Hz. However, to understand the implications of frequency variation on the BiVTC performance, we systematically down-sampled the dataset to $25$Hz, $20$Hz, $15$Hz, and $10$Hz to evaluate the trade-offs between data resolution and computational efficiency at different frequencies. These analysis aims to identify an optimal sampling rate that balances detailed signal representation with the practical constraints of real-time processing. This investigation is crucial to develop a versatile model capable of adapting to various operational scenarios without compromising the accuracy of activity recognition.

%\vspace{1cm}

\begin{figure}[t!]
    \centering
    \includegraphics[width=0.52 \textwidth]{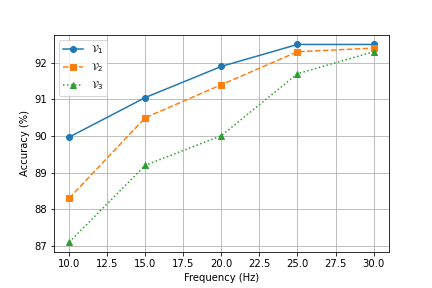}
    \caption{Accuracy of BiVTC at different sampling frequencies in percentage. The model was trained and tested on datasets $\mathcal{V}_1$, $\mathcal{V}_2$, and $\mathcal{V}_3$, corresponding to velocities $v_1$, $v_2$, and $v_3$, respectively, where $v_1 < v_2 < v_3$.}
    \label{fig:freq_bi}
\end{figure}

Figure~\ref{fig:freq_bi} shows the accuracy of BiVTC model for various arm velocities and sampling frequencies. Each data sample encompasses a $12$-second window, with the duration of various robot actions ranging from $2$ to $4$ seconds at velocity $v_1$. This duration corresponds to the acquisition of $60$ to $120$ data packets during active arm movements, with the remaining data capturing periods of silence. When the sampling frequency is reduced to $10$Hz, the number of data packets collected during movement decreases to between $20$ and $40$. This reduction in data volume simplifies the training process but simultaneously raises the risk of overfitting the model. The implications of such overfitting on test accuracy are evident in Figure \ref{fig:freq_bi}. Moreover, an increase in robot velocity shortens the motion period, leading to a smaller amount of data for our model to discern between different classes, which in turn precipitates a significant decline in accuracy.
%\vspace{1cm}

%\vspace{2.5cm}

\subsection{Sniffer Location Selection} \label{sec: loc sec}
Identifying the best sensor placement in robotics is a significant and actively pursued area of research \cite{9992368, vafaee2024real, alali2024deep}. The effectiveness of WiFi sensing for RAR depends on the strategic placement of sniffers, necessitating extensive data collection across various environments to build a model robust to environmental changes, as demonstrated in \cite{widar2019}. This necessity raises the critical question of optimal sniffer location to capture detailed activity data within the environment.

\begin{figure}[t!]
    \centering
    \subfigure[$L_1$]{\includegraphics[width=0.22\textwidth]{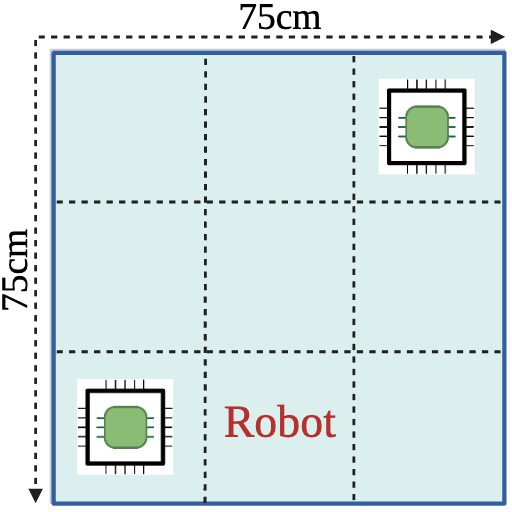}}
    \subfigure[$L_2$]{\includegraphics[width=0.22\textwidth]{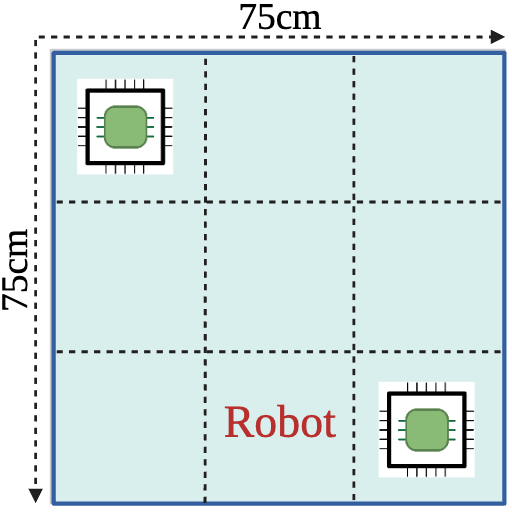}}
    \subfigure[$L_3$]{\includegraphics[width=0.22\textwidth]{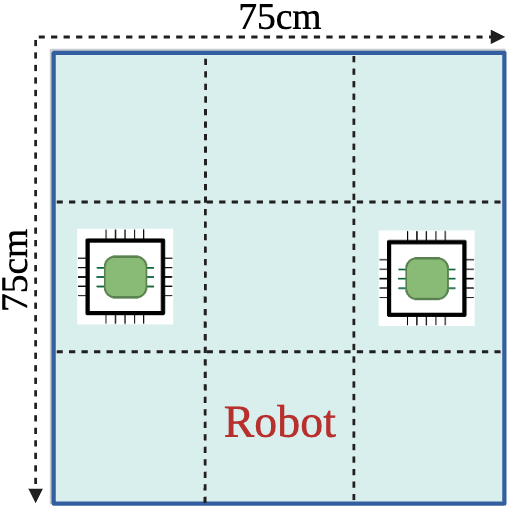}}
    \subfigure[$L_4$]{\includegraphics[width=0.22\textwidth]{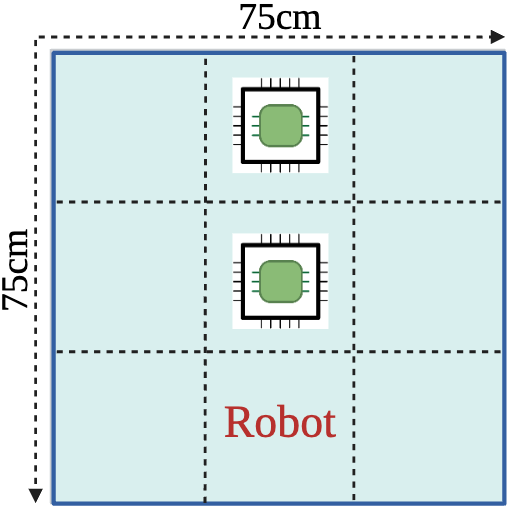}}
    \caption{A 3 x 3 grid illustrating four configurations for sniffer placement, labeled $L_1$, $L_2$, $L_3$, and $L_4$. Each configuration uses eight operational cells, avoiding overlap and maximizing coverage for robust data collection, essential for accurate RAR.}
    \label{fig:locations}
\end{figure}

To investigate the impact of sniffer location for RAR, we constructed a $3 \times 3$ grid, where nine potential locations were identified, with one permanently occupied by the robot's base, leaving eight available for sniffer placement, as presented in Figures~\ref{fig:floor} and~\ref{fig:locations}. Despite having eight locations for two sniffers, where order does not matter, this arrangement allows for $28$ distinct combinations, ensuring that sniffers are not placed too closely (i.e., not in the same grid cell). This setup aims to collect CSI from different locations to enhance model robustness. We adopted a systematic approach to position our sniffers in four distinct areas within the grid, labeled as $L_1, L_2, L_3$, and $L_4$, as depicted in Figure \ref{fig:locations}.

\begin{figure}[t!]
    \centering
    \subfigure[Testing on $L_1$]{\includegraphics[width=0.23\textwidth]{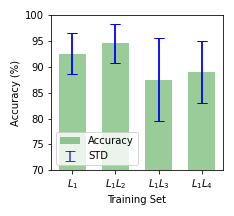}}
    \subfigure[Testing on $L_2$]{\includegraphics[width=0.23\textwidth]{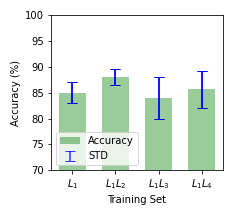}}
    \subfigure[Testing on $L_3$]{\includegraphics[width=0.23\textwidth]{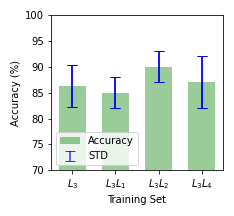}}
    \subfigure[Testing on $L_4$]{\includegraphics[width=0.23\textwidth]{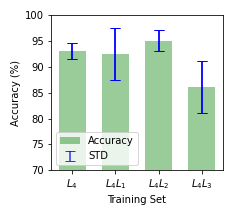}}
    \caption{Performance metrics showing average accuracy and standard deviation across different testing locations, providing insights into the adaptability of the BiVTC model under varying data collection scenarios.}
    \label{fig:l1tr}
\end{figure}

To further our understanding of influence of location, we conducted experiments by deploying the BiVTC model and collected $18$ training and five test CSI samples of each class of robotic arm activity from each designated location, all at the velocity of $v_2$. This data collection method resulted in separate training and testing sets, isolating location as the sole variable, in this study. Initial tests trained and evaluated the BiVTC model in identical locations, revealing accuracy of distinct test sets illustrated in Figure \ref{fig:l1tr}.

\begin{figure}[t!]
    \centering
    \subfigure[Testing on $L_1$]{\includegraphics[width=0.37\textwidth]{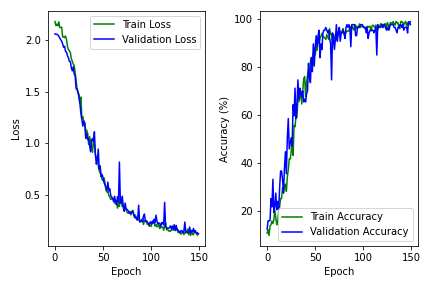}}
    \subfigure[Testing on $L_2$]{\includegraphics[width=0.37\textwidth]{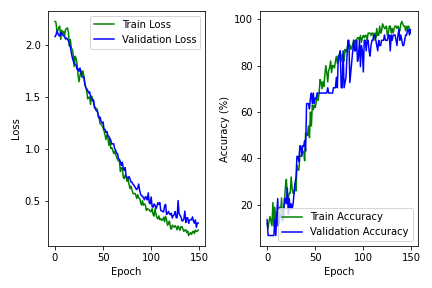}}
    \subfigure[Testing on $L_3$]{\includegraphics[width=0.37\textwidth]{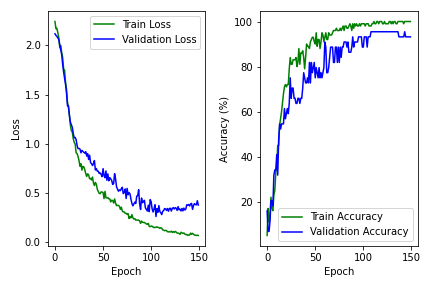}}
    \subfigure[Testing on $L_4$]{\includegraphics[width=0.37\textwidth]{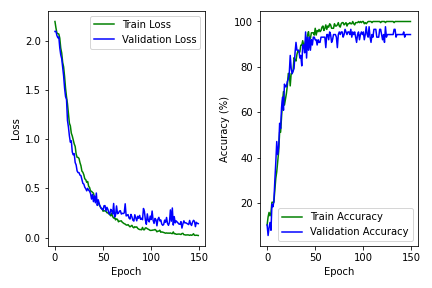}}
    \caption{Training and validation curves of the BiVTC model, tested on different locations. Each figure presents train and validation loss and accuracy, for one distinct location. In this scenario, the model is trained and tested on the data of the same location. }
    \label{fig:trcurve}
\end{figure}

Learning curves for each location were also analyzed to observe training dynamics, as presented in Figure \ref{fig:trcurve}, detailing loss and accuracy trends for training and validation sets. This comprehensive approach aids in determining the optimal sniffer placement for capturing extensive datasets, further allowing us to compare the time and computational resources required to achieve high model accuracy. Our final phase of the study explores the potential of incorporating data from diverse locations as a regularization strategy, aimed at reducing model overfitting and enhancing performance on the test sets.

In location 1 ($L_1$), between the $60^{\text{th}}$ and $70^{\text{th}}$ epochs, the training and validation accuracy averaged $94.3$\% and $92.6$\%, respectively, with the loss dropping to less than $0.5$. By epoch $150$, the training accuracy reached $96.4$\%, and the BiVTC model achieved a test accuracy of $92.5$\%. Notably, the validation loss closely followed the training loss.

In location 2 ($L_2$), the learning curves exhibited a slower slope, indicating that learning data from this location required more time, which can be challenging with limited time and memory resources. This slower learning process also affected the test accuracy, which dropped to $85.0$\%.

In the case of location 3 ($L_3$), there was approximately a $15.0$\% difference between test and validation accuracy, which was more pronounced in the loss plot. This difference was also reflected in the lower test accuracy of $86.2$\%. It is notable, in training our models, we have used early stopping and dropout to prevent overfitting, as discussed in section \ref{sec: Model div}, so the results shown in Figure \ref{fig:trcurve} are only for sake of comparison, and what we observe in between epoch $120^{\text{th}}$ to $150^{\text{th}}$ is prevented the original design of the model.

Finally, in location 4 ($L_4$), which is the closest location to the robotic arm's body, we observed a steep slope in both the validation and training curves. In addition, the test accuracy for $L_4$ increased to $91.5$\%, highlighting the simplicity of this dataset. To augment the dataset, we mixed the training sets and assessed whether adding data from different locations could improve the model's performance in recognizing activities in the test dataset. As shown in Figure \ref{fig:l1tr}, adding data from $L_2$ improved the performance of BiVTC across all locations.

\section{Conclusion \& Future Works}
In autonomous robotics, accurately forecasting robot activities in low-visibility indoor settings is a persistent challenge. Traditional detection and localization methods, which rely on vision or LiDAR,  raise privacy issues and are heavily dependent on LoS for accuracy, often falling short in environments lacking LoS. Our research introduces a novel method using CSI from WiFi signals to precisely identify eight activities of a Franka Emika robotic arm using our BiVTC methodology, irrespective of velocity changes, without additional sensors. We've also released a comprehensive CSI dataset to support further research and collaboration, advancing the use of WiFi signals for robotic action prediction in complex indoor environments.

Looking ahead, our research will focus on advancing machine learning models for more accurate RAR, especially within dynamic environments such as moving objects and people, where traditional methods may not suffice. We aim to broaden our CSI data collection to include dynamic settings, enriching the dataset's diversity and applicability. Additionally, we plan to integrate multimodal data sources, such as acoustic signals and vision sensors, to enhance recognition capabilities, against environmental noises. Exploring real-time RAR systems for reduced latency and improved efficiency in interactive settings is also on our agenda. Moreover, we recognize the importance of addressing the ethical and privacy considerations associated with the deployment of RAR technologies, ensuring their use aligns with societal norms and values.
%\vspace{4cm}

\printbibliography
% \section*{Acknowledgments}
% This should be a 
%\newpage
%\vspace{-2cm}
% {\appendix[Confusion matrices analyses]\label{appendix}
% %\vspace{-1cm}
% As discussed in Section \ref{dif vel}, we assessed the robustness of our models using LOVO cross-validation. According to the results presented in Table \ref{tab:vel classes}, the BiVTC model outperformed the others. To gain a more precise understanding of the performance of this model, we present the confusion matrices for each cross-validation scenario in Figure \ref{fig: conf V}.
% %\vspace{2cm}
% {\appendices
% \section*{}

\end{document}